%% file: main.tex
\pdfoutput=1
\documentclass[11pt]{article}

\usepackage[final]{acl}

\usepackage{times}
\usepackage{latexsym}

\usepackage[T1]{fontenc}
\usepackage[utf8]{inputenc}

\usepackage{microtype}

\usepackage{inconsolata}

\usepackage{graphicx}

\usepackage{xspace}
\usepackage{pifont}

\usepackage{enumitem}
\usepackage{booktabs}
\usepackage{multicol}
\usepackage{multirow}
\usepackage{caption}
\usepackage{subcaption}
\usepackage{adjustbox}
\usepackage{colortbl}
\usepackage[most]{tcolorbox}
\usepackage{makecell}
\usepackage{mdframed,lipsum}
\usepackage{enumitem}
\usepackage[capitalize]{cleveref}

\Crefname{figure}{{Figure}}{{Figures}}
\Crefname{table}{{Table}}{{Tables}}

\setlist[itemize]{leftmargin=*}

\newtcolorbox{mybox}[1]{colback=green!5!white,colframe=green!25!black,fonttitle=\bfseries,title=#1}

\newtcolorbox{myboxcross}[1]{colback=green!5!white,colframe=green!25!black,fonttitle=\bfseries,breakable,title=#1}

\newcommand{\methodname}{\textsc{GuideLLM}\xspace}

\title{\textsc{GuideLLM}: Exploring LLM-Guided Conversation with Applications in Autobiography Interviewing}

\author{
 \textbf{Jinhao Duan\textsuperscript{1}\thanks{ \, Equal contribution.}},
 \textbf{Xinyu Zhao\textsuperscript{2$^*$}},
 \textbf{Zhuoxuan Zhang\textsuperscript{3$^*$}},
 \textbf{Eunhye Ko\textsuperscript{4}},
\\
 \textbf{Lily Boddy\textsuperscript{4}},
 \textbf{Chenan Wang\textsuperscript{1}},
 \textbf{Tianhao Li\textsuperscript{4}},
 \textbf{Alexander Rasgon \textsuperscript{4}},
\\
 \textbf{Junyuan Hong\textsuperscript{4}},
 \textbf{Min Kyung Lee\textsuperscript{4}},
 \textbf{Chenxi Yuan\textsuperscript{5}},
 \textbf{Qi Long\textsuperscript{6}},
\\
 \textbf{Ying Ding\textsuperscript{4}},
 \textbf{Tianlong Chen\textsuperscript{2}},
 \textbf{Kaidi Xu\textsuperscript{1}}
\\
\\
 \textsuperscript{1}Drexel University,
 \textsuperscript{2}UNC Chapel Hill,
 \textsuperscript{3}Brown University,
 \textsuperscript{4}UT Austin,
 \\
 \textsuperscript{5}New Jersey Institute of Technology,
 \textsuperscript{6}University of Pennsylvania
\\
 \small{
   \textbf{Correspondence:} \href{mailto:email@domain}{kx46@drexel.edu}
 }
}

\begin{document}
\maketitle

\input{sections/abstract}

\input{sections/section1_introduction}

\input{sections/section2_related_work}

\input{sections/section3_methodology}

\input{sections/section4_experiments}

\input{sections/human_involved_results}

\input{sections/section5_conclusion}

\bibliography{custom}

\input{sections/appendix}

\end{document}

%% file: sections/abstract.tex
\begin{abstract}
Although Large Language Models (LLMs) succeed in human-guided conversations such as instruction following and question answering, the potential of LLM-guided conversations—where LLMs direct the discourse and steer the conversation's objectives—remains under-explored. In this study, we first characterize LLM-guided conversation into three fundamental components: (\textit{i}) \textit{Goal Navigation}; (\textit{ii}) \textit{Context Management}; (\textit{iii}) \textit{Empathetic Engagement}, and propose \methodname as an installation. We then implement an interviewing environment for the evaluation of LLM-guided conversation.
Specifically, various topics are involved in this environment for comprehensive interviewing evaluation, resulting in around 1.4k turns of utterances, 184k tokens, and over 200 events mentioned during the interviewing for each chatbot evaluation.
We compare \methodname with 6 state-of-the-art LLMs such as GPT-4o and Llama-3-70b-Instruct, from the perspective of interviewing quality, and autobiography generation quality. For automatic evaluation, we derive user proxies from multiple autobiographies and employ LLM-as-a-judge to score LLM behaviors. We further conduct a human-involved experiment by employing 45 human participants to chat with \methodname and baselines. We then collect human feedback, preferences, and ratings regarding the qualities of conversation and autobiography. Experimental results indicate that \methodname significantly outperforms baseline LLMs in automatic evaluation and achieves consistent leading performances in human ratings.

\end{abstract}

%% file: sections/section1_introduction.tex
\section{Introduction}
\begin{figure*}[t]
    \centering
    \includegraphics[width=1\linewidth]{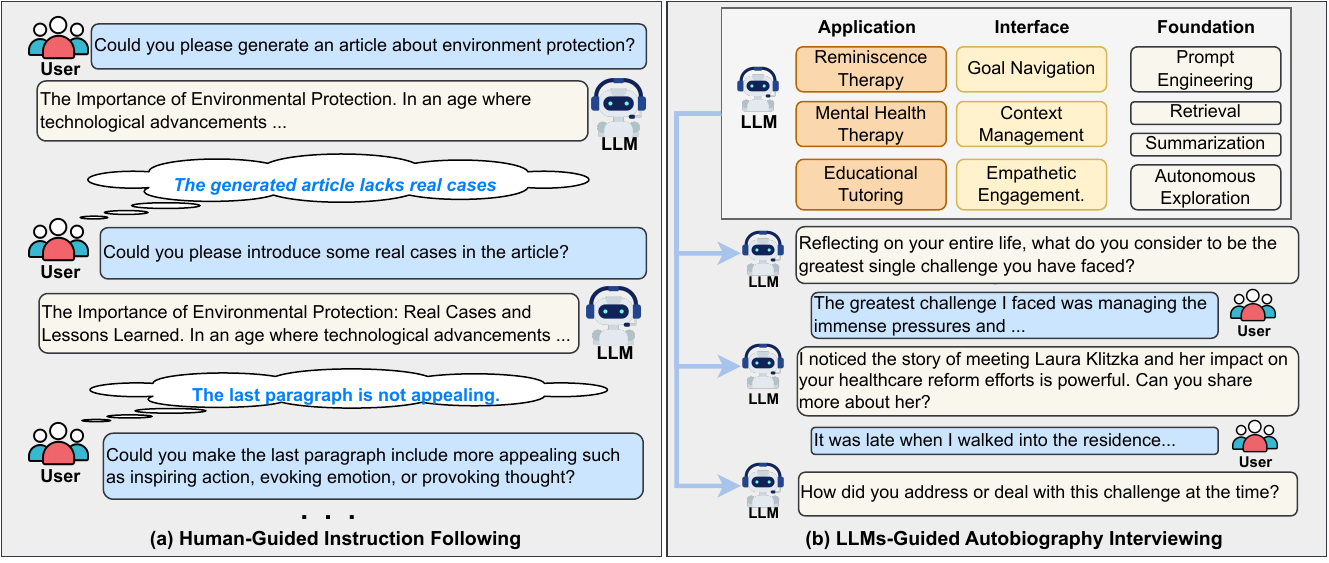}
    \caption{Comparison between human-guided conversation and LLM-guided conversation. (a) Human-Guided: Human dominates the conversation, providing feedback and instruction to LLMs. (b) LLM-Guided: LLMs navigate the goal by automatically extrapolating interview questions.}
    \label{fig:human_guided_vs_llm_guided}
    \vspace{-5mm}
\end{figure*}

Large Language Models (LLMs) have demonstrated their effectiveness in \textit{human-guided} dialogue, in which LLMs are tasked with producing responses according to specific commands from human operators, such as instruction following~\cite{ouyang2022training} and question answering~\cite{chang2024survey}. In this type of task, the primary duty of  LLMs is to adhere to the instructions given by humans to ensure the generated output is accurate and close to human expectations, as shown in~\cref{fig:human_guided_vs_llm_guided}(a).

However, tasks in the real world are more complex, necessitating greater autonomy from LLMs~\cite{wang2024survey,duan2022survey,wu2023autogen}. For example, tasks such as interviewing are dramatically different from traditional tasks as interviewing is open-ended, without definitive or ``perfect'' outcomes. Interviewing tasks demand that LLMs plan the interview procedure, manage the objectives, e.g., \textit{exploring the user's memory and life experiences in autobiography interviewing}, and offer adaptive and personalized inquiries based on the users' responses. This conversation paradigm requiring LLMs to guide and manage the conversation, ensuring the conversation flows smoothly and the objectives are met, is termed as \textit{LLM-guided} conversation (~\cref{fig:human_guided_vs_llm_guided}(b)).

There have been related works in the LLM-guided conversation, such as role-play~\cite{wang2024incharacter,wang2023rolellm,chen2024roleinteract,tao2023rolecraft,li2023chatharuhi} and goal-oriented LLMs~\cite{ham-etal-2020-end,NEURIPS2020_e9462095,wu-etal-2020-tod,MehriDialoGLUE2020,Inagaki2023LLMsCG}. For role-play LLMs, they either prompt LLMs to perform specific roles such as a patient~\cite{wang2024patient}, doctor~\cite{panagoulias2024augmenting}, gamer~\cite{duan-etal-2024-reta, duan2024gtbench}, or investigate the human-like features of LLMs, e.g., emotions~\cite{li2023large} and personalities~\cite{safdari2023personality}. 
Goal-oriented LLMs enable the model to attain greater levels of autonomy, particularly in fields such as space exploration~\cite{maranto2024llmsat}.
While role-play and goal-oriented LLMs provide some autonomy and allow for the simulation of a specific role, their ability to actively control and effectively handle a full conversation is still underexplored.

In this paper, we investigate the LLM-guided conversation from framework design to autobiography interviewing applications. 
Inspired by popular social science theories, PEACE Model~\cite{clarke2001national} and Motivational Interviewing (MI)~\cite{hettema2005motivational}, we design
\methodname to comply with these models by comprising three pivotal components: (\textit{i}) \textbf{Goal Navigation} module, as the cornerstone of the framework, steers the conversation with pre-defined interviewing protocols and dynamic memory graphs for extrapolating dialogue trajectories. (\textit{ii}) \textbf{Context Management} module iteratively distills the main idea of each session into a contextual summary for subsequent sessions.
(\textit{iii}) \textbf{Empathetic Engagement} module refines LLM response with expression strategies by the real-time monitoring of user emotion. 

For evaluation, we create an interviewing environment where \methodname is tasked with conducting interviews over 23 different topics, ultimately producing an autobiography based on the outcomes of these interviews. Then, the behaviors of \methodname and baselines are evaluated in three-folds: (\textit{i}) {Interviewing Quality}, {e.g., event coverage and correctness}; (\textit{ii}) {Conversation Quality} {e.g., communication fluency, identification, and comforting}; (\textit{iii}) {Generation Quality}, {e.g., the insightfulness, narrativity, and emotional impact of the generated autobiography}. We also carry out human-involved experiments with 45 participants, prompting them to engage in conversations with \methodname and baseline models. Following these interactions, we gather feedback, preferences, and ratings from the participants.
Our contributions can be summarized as the following:
\begin{itemize}

    \item \textbf{Framework.} We define the realm of LLM-guided conversations and propose \methodname as an installation within this conversational paradigm. There are three components comprised in \methodname: Goal Navigation, Context Management, and Empathetic Engagement.

    \item \textbf{Technique.} \methodname effectively harnesses a variety of techniques such as Retrieval Augmented Generation (RAG) and long-context summarization to boost the ability of LLMs to effectively lead and steer a conversation. Moreover, a memory graph is designed to drive memory extrapolation, thereby enhancing the goal navigation capabilities of \methodname.

    \item \textbf{Application.} We present the autobiography interviewing environment as a practical application of LLM-guided conversations. Within this setting, LLMs are tasked with initiating and steering the interview with users, aiming to generate a comprehensive autobiography.

    \item \textbf{Evaluation.} We propose a comprehensive evaluation protocol for our LLM-guided autobiography interviewing environment, including interview quality, conversation quality, and autobiography generation evaluation, encompassing both LLM-as-a-judge evaluation and human subjects evaluations.
\end{itemize}

%% file: sections/section2_related_work.tex
\section{Related Work}

\noindent\textbf{Role-Play LLM.} Role-playing agents (RPAs) powered by large language models (LLMs) are challenged by the evaluation of fidelity to target personas. Traditional methods focus on replicating characters' knowledge and linguistic patterns, requiring character-specific datasets. ~\citet{huang2023psychobench} evaluate LLM personalities with self-report scales (BFI and MBTI), targeting LLM psychometric properties but not specifically addressing persona adherence.~\citet{li2023chatharuhi} and~\citet{wang2023rolellm} develop character-specific RPAs to enhance conversational abilities, human-likeness, and multi-turn consistency. 
However, they have not deeply explored character fidelity. 
\citet{tao2024chatgpt} found that adapting responses based on emotional cues significantly improved user satisfaction in role-playing scenarios. \citet{tao2023rolecraft} also demonstrated that with a scalable and controlled learning environment, LLM-driven simulations could effectively mimic real-life interactions.

\noindent\textbf{Long Text Generation and Management.}
The key techniques for long text generation and management considered in this study include summarization and Retrieval Augmented Generation (RAG).
\citet{luo2023chatgpt} explore using ChatGPT to evaluate factual consistency in summarization. \citet{zhong2022dialoglm} propose a pre-training framework for long dialogue understanding and summarization using a window-based denoising approach. \citet{xu2022sequence} introduce a contrastive learning model, SeqCo, to improve the faithfulness of abstractive text summarization. \citet{zhang2021summ} efficiently processes long texts by dividing them into manageable segments and summarizing each iteratively.
\citet{gao2023retrieval} overview RAG, which integrates external knowledge from databases to enhance LLM-generated content's reliability.

%% file: sections/section3_methodology.tex
\section{LLM-Guided Autobiography Interview}
In this section, we first introduce LLM-guided conversation and its general framework. Then, we introduce \methodname as an implementation of LLM-guided conversation, centering on interviewing.

\begin{figure*}[t]
    \centering
    \includegraphics[width=1\textwidth]{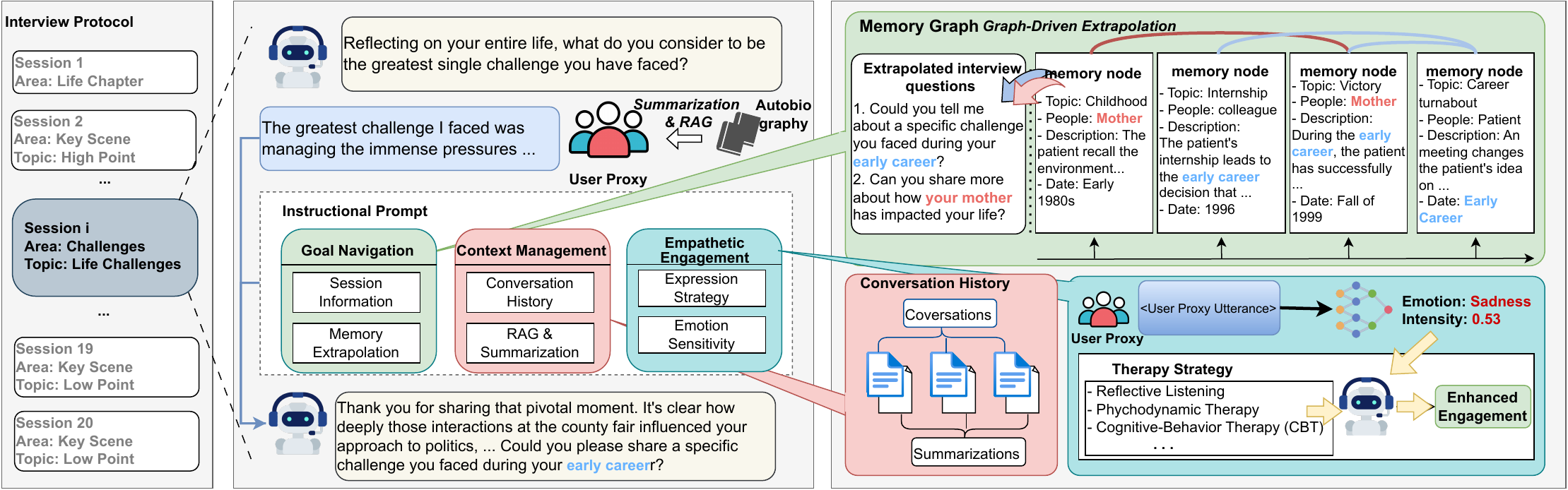}
    \caption{The overall pipeline of \methodname in the guided conversation environment.}
    \label{fig:llm_as_autobiographer}
    \vspace{-5mm}
\end{figure*}

\subsection{LLM-Guided Conversation}
Several theories from social science research provide comprehensive protocols for how to conduct guided conversations, e.g., interviewing. For instance, the PEACE Model~\cite{clarke2001national} and Motivational Interviewing (MI)~\cite{hettema2005motivational} highlight ``\textit{engage and explain}'' and ``\textit{planning and preparation}''. To comply with these theories, \methodname is designed from three essential qualities that an LLM should possess for effective conversation guidance:

\paragraph{\textit{Goal Navigation.}} LLM steers the conversation and determines pivotal transitions, initiatively exploring and extrapolating new components that can shift the conversation toward the intended outcome.
\paragraph{\textit{Context Management.}} LLMs summarize the ongoing dialogue, resuming previous discussions, connecting current conversations with past ones, and managing historical data exchanged between the LLM and users.
\paragraph{\textit{Empathetic Engagement.}} LLMs interact with the user by providing empathetic responses, delivering suitable tone and content, and demonstrating sensitivity towards the user's emotional state.

LLM-guided conversations extend to numerous practical applications. For instance, in {\textit{Interviews}}, they can frame pertinent questions, guiding the conversation. In {\textit{Educational Tutoring}}~\cite{nye2023generative}, LLM-guided conversations can assess the user's state, crafting specific and personalized plans to facilitate the learning process. Another promising application is {\textit{Mental Health Therapy}}~\cite{demszky2023using,hong2024conect}, where the LLM can guide the conversation, applying personalized therapy based on the user's responses. \cref{fig:llm_as_autobiographer} provides a overall pipeline for \methodname in guided conversations. We will use the autobiography interviewing task as an example of guided conversations for the rest of this paper.

\subsection{Goal Navigation}
In guided conversations, LLMs are responsible for guiding the conversation, delivering adaptive responses to users, and ensuring that the conversation objectives are met.
To accomplish this, we utilize a hybrid approach, combining \textit{Verbalized Interviewing Protocol (VIP)}~\cite{2016maunsell,castillo2016preparing,lamb2007structured} with \textit{Memory Graph-Driven Extrapolation (MGE)}:

\paragraph{Verbalized Interviewing Protocol (VIP).} We leverage the popular interviewing protocol, \textit{``The Life Story Interview''}~\cite{mcadams2008life}, as the general guidance. 
This protocol covers essential topics including \underline{\textit{Life Chapters}}, \underline{\textit{Key Scene in Life}}, \underline{\textit{Future}}, \underline{\textit{Challenges}}, and \underline{\textit{Personal Ideology}}. For instance, in the \textit{Challenges} area, topics such as \textit{Life Challenges}, \textit{Health}, \textit{Loss}, and \textit{Regrets} will be covered. We design system prompts according to specific topics, containing basic seed questions, to make LLMs primarily focus on one topic during each session. In this way, the entire interviewing process could be structured into 23 sessions of conversations between the user and the LLM chatbot. 
Please refer to~\cref{appendix:interview_protocol} for more details of the interviewing protocol and the prompt templates.

\paragraph{Memory Graph Extrapolation (MGE).}\label{sec:memory_graph}
The objective of MGE is to explore unique characteristics and generate adaptive questions for personalized interviewing. MGE operates as an LLM-driven function, performing various operations on memories, including (i) \textit{extracting} events from conversations, (ii) \textit{inserting} new events into the graph, (iii) \textit{merging} existing events, and (iv) \textit{extrapolating} queries based on current events. Each event is associated with properties such as Date, People Involved, and Event Description.

In essence, MGE follows an ``extract, merge, then extrapolate'' process: initially, memory is initialized based on the user utterance. Next, the extraction process identifies and lists any events mentioned by the user, which are then merged with the existing events. Finally, MGE generates personalized questions from the event nodes by uncovering various relationships, such as identifying individuals who are mentioned frequently. These questions are stored in a "question cache" for use in the subsequent turn of the conversation. For details on the prompt templates used for event extraction and memory extrapolation, please refer to~\cref{appendix:mge}.

\subsection{Context Management}
Context is a crucial information source for long conversations where each autobiographer has a unique and lengthy personal experience.
However, LLMs have limited context lengths that can be easily exceeded when processing autobiographies~\cite{dai2019transformer}. Besides, long-context input may bring performance decrease~\cite{Liu2023LostIT} and huge financial cost for the close-sourced. To tackle this, we incorporate a context management module in our framework capable of progressively summarizing and retrieving conversation history.

\paragraph{Conversation History Summarizing.}
Inspired by~\citeauthor{Chang2023BooookScoreAS,maharana2024lococmo}, we implement an iterative summarization process that generates a summary for the current session based on the summaries from previous sessions, providing additional context for the chatbot. When initializing the chatbot, if a history conversation file is present in the configuration, we first generate a summary of the loaded conversation. At the start of each conversation after the first, the system prompt of the LLM chatbot includes a summarization section (see Appendix~\ref{appendix:system_prompt}), beginning with an instruction indicating that it has previously conversed with the user, followed by the specific summary of the prior session. The prompt and summarization pipeline are detailed in Appendix~\ref{appendix:summary_prompt} and Figure~\ref{fig:summary_pipe}.

\subsection{Empathetic Engagement}\label{sec:empathetic_engagement}
LLM-guided conversations should accurately understand the user's state and respond appropriately. This involves empathetic interaction, creating a space where users feel at ease to share more about themselves.  We accomplish this by enhancing \textit{Expression Strategies} and \textit{Emotion Detection}:

\input{tables/memory_exploration}
\input{tables/jane_eyre}

\paragraph{Expression Strategy.} To enhance the expression capability of LLMs, we draw inspiration from popular mental health therapy strategies, including Reflective Listening~\cite{rautalinko2007reflective}, Cognitive-Behavior Therapy (CBT)~\cite{beck2020cognitive}, and Psychodynamic Therapy~\cite{leichsenring2003effectiveness}. Although originally designed to address mental health issues, these therapeutic strategies offer insightful guidance on effective communication with users and provide meaningful advice on interaction techniques. The introduction to therapy strategy and prompts are in~\cref{appendix:therapy_strategy}.

\paragraph{Emotion Detection.} Emotion sensitivity is a critical element in conveying the state of individuals that has been significantly underscored in human-computer interaction~\cite{cowie2001emotion, brave2007emotion} and LLMs~\cite{li2023large}. 
To enhance the emotional sensitivity of LLMs, we employ EmoLlama-7b~\cite{liu2024emollms} for the emotion detection of user utterances. Specifically, we prompt EmoLlama to provide both the emotion category (one of the emotions including \textit{anger, anticipation, disgust, fear, joy, love, optimism, pessimism, sadness, surprise, trust}) and its intensity (\textit{from 0 to 1}) for user response. We then guide the LLM to generate suitable responses that align well with the user's emotional state, e.g., including expressions of empathy or comfort when detecting an upset user. Please refer to~\cref{appendix:emotion} for more details.

\subsection{Autobiography Generation}
\input{tables/statistics}

Autobiography holds a distinctive form in comparison to other book categories, as an autobiography typically consists of numerous individual chapters, each of which relays a specific spirit or theme intimately tied to the author's life. This format aligns seamlessly with our structured interview protocol; the scope and topics encompassed in the interview protocols are similarly singular and targeted, allowing for a thorough exploration of each subject.

Therefore, when generating an autobiography, we generate each 
chapter by sequentially building upon each interviewing session.
Specifically, for each session, we meld the conversation history and memory nodes derived from the current session, then prompt GPT-4 to emphasize the key areas and topics discussed in that particular session.
Please refer to~\cref{appendix:autobiography_generation} for more details of autobiography generation.

%% file: tables/memory_exploration.tex
\begin{table}[t]
    \centering
    \adjustbox{width=\linewidth}{
    \begin{tabular}{lcccc}
        \toprule
        &  & \multicolumn{3}{c}{\textbf{Correctness} (\%)} \\
        \cmidrule(lr){3-5}
        \multicolumn{1}{c}{\textbf{Model}} & \textbf{$coverage$} & \textbf{$P.$ }& \textbf{$Recall$} & \textbf{$F1$}  \\
        \toprule
        \multicolumn{5}{c}{\textit{``A Promised Land''}} \\
        \midrule
        GPT-4-turbo & 42.8 & 17.0 & 5.8 & 4.3 \\
        GPT-4o & 57.1 & 22.0 & 7.9 & 5.8 \\
        Llama-3-70b-Instruct & 57.1 & 22.4 & 14.3 & 8.7 \\
        Mixtral-8x22B-Instruct-v0.1 & 28.5 & 13.3 & 4.3 & 3.2 \\
        Qwen2-72b-Instruct & 28.6 & 11.9 & 3.5 & 2.7 \\
        \midrule
        \rowcolor[gray]{0.9}
        \methodname (ours) & \textbf{85.7} & \textbf{69.4} & \textbf{47.4} & \textbf{28.2} \\
        \midrule
        \multicolumn{5}{c}{\textit{``An Autobiography by Catherine Helen Spence''}} \\
        \midrule
        GPT-4-turbo & 21.0 & 40.0 & 20.1 & 13.4 \\
        GPT-4o & 5.2 & 21.5 & 14.2 & 8.5 \\
        Llama-3-70b-Instruct & 0.0 & 23.4 & 12.6 & 8.1 \\
        Mixtral-8x22B-Instruct-v0.1 & 0.0 & 34.1 & 11.7 & 8.7 \\
        Qwen2-72b-Instruct & 5.3 & 28.2 & 10.9 & 7.8 \\
        \midrule
        \rowcolor[gray]{0.9}
        \methodname (ours) & \textbf{36.8} & \textbf{68.3} & \textbf{68.9} & \textbf{34.3} \\
        \bottomrule
    \end{tabular}
    }
    
    \caption{Interviewing quality evaluation. $P.$ stands for $Precision$.}
    \label{tab:memory_explore}
    \vspace{-3mm}
\end{table}

%% file: tables/jane_eyre.tex
\begin{table}[h]
    \centering
    \adjustbox{width=0.9\linewidth}{
    \begin{tabular}{lccc}
        \toprule
        &  \multicolumn{3}{c}{\textbf{Correctness} (\%)} \\
        \cmidrule(lr){2-4}
        \multicolumn{1}{c}{\textbf{Model}} & \textbf{$Precision$ }& \textbf{$Recall$} & \textbf{$F1$}  \\
        \toprule
        GPT-4-turbo &  20.3 & 6.9 & 5.1 \\
        GPT-4o & 13.6 & 5.6 & 4.0 \\
        Llama-3-70b-Instruct & 13.6 & 7.5 & 4.8 \\
        Mixtral-8x22B-Instruct-v0.1 & 23.0 & 5.6 & 4.5 \\
        Qwen2-72b-Instruct & 25.0 & 4.4 & 3.7 \\
        \midrule
        \rowcolor[gray]{0.9}
        \methodname (ours) & \textbf{73.4} & \textbf{22.6} & \textbf{17.3} \\
        \bottomrule
    \end{tabular}
    \vspace{-2mm}
    }
    
    \caption{Interviewing quality evaluation on \textit{``Jane Eyre: An Autobiography''}. }
    \label{tab:memory_explore_janeeyre}
    \vspace{-8mm}
\end{table}

%% file: tables/statistics.tex
\begin{table}[h]
    \centering
    \adjustbox{width=\linewidth}{
    \begin{tabular}{cccccc}
    \toprule
 \textbf{Model}   & 
 \textbf{Topics}    & 
 \makecell{\textbf{Turns}\\\textbf{in total}}   & 
 \makecell{\textbf{Tokens}\\\textbf{per conv.}\\\textbf{(Avg.)}} & 
 \textbf{Event} & 
 \makecell{\textbf{Event}\\\textbf{token}\\\textbf{usage (Avg.)}} \\ \midrule
        (Avg.) baselines    & 23 & 1380  & 8,443        & 206     & 1,229.56       \\
\methodname & 23 & 1380  & 3,932        & 262     & 900.46     \\ \toprule
\end{tabular}
    \vspace{-3mm}
    }
    \caption{The conversation statistics in our interviewing environment. The tokens are counted by the tokenizer of GPT-4. ``Conv.''=Conversation. ``Event token usage (Avg.)'' reflects the efficiency of extracting events from dialogues, lower means more efficient conversation.}
    \label{tab:stats}
    \vspace{-5mm}
\end{table}

%% file: sections/section4_experiments.tex
\section{Experiments for Automatic Evaluation}\label{sec:automatic_eval_exp}
\input{tables/merged_conv_autobio}

\subsection{Experimental Settings}
\noindent\textbf{User Proxy.} We utilize GPT-4-turbo to simulate users for evaluating \methodname in autobiography interviewing. Three LLM user proxies are implemented based on popular autobiographies: \textit{``A Promised Land''} by Barack Obama, \textit{``Jane Eyre: An Autobiography''}, and \textit{``An Autobiography by Catherine Helen Spence''}. They are assigned to role-play the corresponding main character in the autobiography and respond to questions by referencing the autobiography using the Retrieval-Augmented Generation (RAG) approach. See~\cref{appendix:user_proxy} for more details.

\noindent\textbf{Evaluation.} Each chatbot will engage in conversations with all user proxies across 23 interview topics (\cref{appendix:interview_protocol}). The evaluation is three-fold:

\noindent\textbf{(1) Interviewing Quality} measures the capability of LLMs to explore users' major events and life experiences and their ability to document these experiences accurately (\cref{section:interviewing_eval}).
    
\noindent\textbf{(2) Conversation Quality} evaluates whether the responses from the LLM chatbot are comforting and engaging (\cref{section:conversation_eval}).
    
\noindent\textbf{(3) Autobiography Generation:} measures the quality of the generated autobiography, such as insightfulness and narrativity (\cref{section:autobio_eval})

\noindent\textbf{Baseline.} To evaluate the design of \methodname, we employ state-of-the-art LLMs and prompt them to be autobiography interviewers.
For a fair comparison, baseline agents are also equipped with basic goal navigation and context management functions. Please refer to~\cref{appendix:baseline_prompts} for how baseline agents are built. The backbone LLMs for \methodname is fixed to GPT-4o, with the same generative hyperparameters as the baselines. We consider both commercial LLMs, e.g., GPT-4 and GPT-4o~\cite{achiam2023gpt}, and open-source LLMs, e.g., Llama-3-70b-Instruct~\cite{llama3}, Mixtral-8x22B-Instruct~\cite{jiang2024mixtral}, and Qwen2-72b-Instruct~\cite{bai2023qwen}.

\subsection{Interviewing Quality Evaluation}\label{section:interviewing_eval}
We denote by $E_{intw}=\{e_1, e_2, \cdots\}$ the events extracted during interviewing conversation. 
We denote by $E_{GT}=\{e_1, e_2, \cdots\}$ the events directly extracted from the original. For both \methodname and baselines, $E_{intw}$ are obtained by prompting LLMs to extract events from conversation history (please refer to ~\cref{appendix:interviewing_coverage} for more details):

\noindent\textbf{Interviewing Coverage ($coverage$)} is calculated by the \underline{\textit{date-intersection}} between $E_{intw}$ and $E_{GT}$:

    \small
    $$
    coverage = \frac{|E_{intw} \cap E_{GT}|}{|E_{GT}|} \times 100\%,
    $$
    \normalsize
    where $e_i \in E_{intw} \cap E_{GT} $ if $e_i \in E_{intw}$ and $\exists \,\, e_j \in E_{GT}$ that has the same date as $e_i$, and $|\cdot|$ is the number of elements. 100\% $coverage$ indicates that all the important dates in the user's life are at least mentioned during the interview.
    
\begin{figure*}[t]
    \centering
    \includegraphics[width=0.9\textwidth]{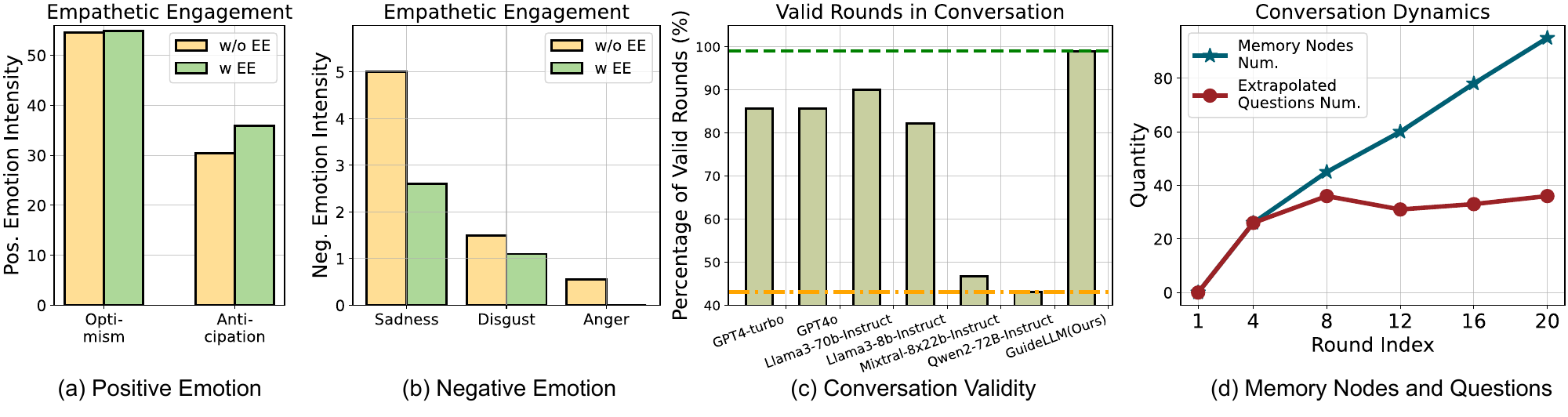}
    \vspace{-3mm}
    \caption{Ablation study of \methodname: how empathetic engagement affects users' (a) positive and (b) negative emotional distributions, (c) statistical results on the number of valid conversation rounds, and (d) the benefits of the MGE in goal navigation.}
    \label{fig:ablation_study}
    \vspace{-5mm}
\end{figure*}

\noindent\textbf{Correctness.} We define the $Precision$ as the percentage of extracted events that are being verified as correct:
    \small
    $$
    Precision = \frac{|E_{correct}|}{|E_{intw}|} \times 100\%, \,\, E_{correct} \subset E_{intw} 
    $$
    \normalsize
    Please refer to~\cref{appendix:interviewing_coverage} for the definition of a correct event. 
    For comprehensive evaluation, we also adopt $Recall$ and $F1$ as evaluation metrics.

 The results of Interviewing Coverage and Correctness are presented in~\cref{tab:memory_explore,tab:memory_explore_janeeyre}, where \methodname shows a significant advantage in Interviewing Coverage over the baselines. This indicates that VIP and MGE are effective techniques for the goal navigation of LLMs in the interviewing framework. For Correctness, \methodname also outperforms baseline agents. Moreover, the $Recall$ and $F1$ indicate that \methodname not only maintains high accuracy for documentation but also extracts more memory events than baselines. The statistics of the interviewing environment are summarized in~\cref{tab:stats}. 

\subsection{Conversation Quality}\label{section:conversation_eval}

Inspired by human evaluation metrics of therapy chatbots from~\citet{Wang2023EnhancingTC}, we design three conversation quality metrics: (\textit{i}) \textbf{\textit{Fluency}}; (\textit{ii}) \textbf{\textit{Identification}}; (\textit{iii}) \textbf{\textit{Comforting}}. We utilize the popular LLM-as-a-judge~\cite{zheng2024judging} evaluation to make GPT-4 decide which conversation is better. Then, we calculate the \textbf{win rate (WR)} and \textbf{loss rate (LR)} of \methodname against baselines. See~\ref{appendix:conv_eval} for the evaluation protocol and prompt templates.

The results in~\cref{tab:conversation_eval} show that \methodname significantly outperforms most baselines in GPT-4-as-a-judge evaluations. With human examinations, we find that baseline agents often resort to simple greetings or summaries, e.g., \textit{``Your commitment to sharing experiences and insights that inspire action and change is truly admirable.}''. Instead of proficiently steering the dialogue to complete the interview, these \underline{\textit{repetitive utterance}} happen multiple times in a session with a baseline agent (\cref{appendix:invalid_repetitive_conv}). In contrast, with our goal navigation module, \methodname provides substantial content at each round of conversation.

\subsection{Autobiography Generation Evaluation}\label{section:autobio_eval}
We follow popular memo evaluations from~\citet{quoraRateCritique, 10.1093/actrade/9780199669240.001.0001, jbp:/content/journals/10.1075/ni.21.2.08smo, 10.1159/000100939} and design three metrics of generated autobiography: \textbf{\textit{Insightfulness}}, \textbf{\textit{Narrativity}}, and \textbf{\textit{Emotional Impact}} (prompt templates can be found in~\cref{appendix:autobio_eval}). Leveraging the same LLM-as-a-judge evaluation protocol in~\cref{section:conversation_eval}, we found that the autobiography generated by \methodname is more favorable than that of baseline agents. 
Examples of generated autobiography are presented in~\cref{appendix:autobio_examples}. 

\subsection{Ablation Study}

\noindent\textbf{Empathetic Engagement.} We study how the Empathetic Engagement (EE) module (\cref{sec:empathetic_engagement}) affects the emotion distribution of user responses. We compare how the intensity of emotions (both positive emotions~\cref{fig:ablation_study}(a) and negative emotions~\cref{fig:ablation_study}(b)) changed when the EE module is enabled and disabled. It is shown that the EE module effectively enhances the user's positive emotions while mitigating negative emotions, indicating that express strategy and emotional sensitivity foster a more positive emotion for users.

\noindent\textbf{Valid Rounds in Conversation.} As outlined in~\cref{section:conversation_eval}, the lack of autonomy in LLMs leads to repetitive responses. We manually count 10 conversation sessions for each chatbot and identify those conversations become repetitive or diverge into irrelevant or nonsensical content. The valid round percentages are calculated as $\frac{\#total\,\,rounds - \#invalid\,\,rounds}{\#total\,\,rounds}$. As shown in~\cref{fig:ablation_study} (c), all baseline models, especially Qwen2-72b-Instruct, show over 50\% meaningless repeats. However, \methodname with a goal navigation module offering diverse and detailed interview questions, has an extensive range of topics and is less prone to repetition.

\noindent\textbf{Conversation Dynamics.} In~\cref{fig:ablation_study} (d), we count the memory events extracted and questions extrapolated at different conversation rounds. Generally, the MGE module identifies around 100 events and extrapolates nearly 40 questions for the LLM's follow-up. This highlights MGE's effectiveness in event management and goal navigation.

%% file: tables/merged_conv_autobio.tex
\begin{table*}[t]
    \centering
    \adjustbox{width=\textwidth}{
    \begin{tabular}{cl|cc|cc|cc|cc|cc|cc}
        \toprule
        & & \multicolumn{6}{c|}{\textbf{Conversation Quality}} & \multicolumn{6}{c}{\textbf{Autobiography Quality}} \\
        \midrule
         \multicolumn{2}{c|}{\textbf{LLM-as-a-Judge}}& \multicolumn{2}{c|}{\textbf{Fluency}} & \multicolumn{2}{c|}{\textbf{Identification}} & \multicolumn{2}{c|}{\textbf{Comforting}} & \multicolumn{2}{c|}{\textbf{Insightfuness}} & \multicolumn{2}{c|}{\textbf{Narrativity}} & \multicolumn{2}{c|}{\makecell{\textbf{Emotional} \\ \textbf{Impact}}}\\
         \cmidrule{1-2}
         \cmidrule(lr){3-4}
         \cmidrule(lr){5-6}
         \cmidrule(lr){7-8}
         \cmidrule(lr){9-10}
         \cmidrule(lr){11-12}
         \cmidrule(lr){13-14}
        \textbf{Ours} & \multicolumn{1}{c|}{\textbf{Baselines}} & WR & LR & WR & LR & WR & LR & WR & LR & WR & LR & WR & LR \\
        \toprule
        \multicolumn{14}{c}{\textit{``A Promised Land''}} \\
        \midrule
        \multirow{6}{*}{\makecell{\methodname (ours) \\ v.s.}} & GPT-4-turbo &\cellcolor{cyan!20}  \textbf{35} & 25 & 50 & 50 & \cellcolor{cyan!20} \textbf{90} & 10 &\cellcolor{cyan!20} \textbf{80} & 20 &\cellcolor{cyan!20} \textbf{90} & 10 & \cellcolor{cyan!20}\textbf{95} & 5 \\
        &GPT-4o &\cellcolor{cyan!20} \textbf{80} & 0 &\cellcolor{cyan!20} \textbf{65} & 35 &\cellcolor{cyan!20} \textbf{95} & 5 & \cellcolor{cyan!20}\textbf{100} & 0 &\cellcolor{cyan!20} \textbf{100} & 0 &\cellcolor{cyan!20} \textbf{85} & 15 \\
        &Llama-3-70b-Instruct &\cellcolor{cyan!20} \textbf{80} & 10 &\cellcolor{cyan!20} \textbf{55} & 40 & 35 & \textbf{65} & \cellcolor{cyan!20}\textbf{75} & 20 &\cellcolor{cyan!20} \textbf{75} & 20 & 45 & \textbf{55} \\
        &Llama-3-8b-Instruct & \cellcolor{cyan!20} \textbf{85} & 10 & \cellcolor{cyan!20}\textbf{65} & 35 &\cellcolor{cyan!20} \textbf{100} & 0 &\cellcolor{cyan!20} \textbf{100} & 0 & \cellcolor{cyan!20}\textbf{100} & 0 & \cellcolor{cyan!20}\textbf{60} & 40\\
        &Mixtral-8x22B-Instruct-v0.1 &\cellcolor{cyan!20} \textbf{100} & 0 & \cellcolor{cyan!20}\textbf{100} & 0 &\cellcolor{cyan!20} \textbf{100} & 0 &\cellcolor{cyan!20} \textbf{100} & 0 &\cellcolor{cyan!20} \textbf{100} & 0 &\cellcolor{cyan!20} \textbf{100} & 0\\
        &Qwen2-72b-Instruct &\cellcolor{cyan!20} \textbf{90} & 10 & \cellcolor{cyan!20}\textbf{85} & 15 &\cellcolor{cyan!20} \textbf{95}& 5  &\cellcolor{cyan!20} \textbf{95} & 0 & \cellcolor{cyan!20}\textbf{95} & 5 &\cellcolor{cyan!20} \textbf{85}& 15 \\
        \midrule
        \multicolumn{14}{c}{\textit{``An Autobiography by Catherine Helen Spence''}} \\
        \midrule
        \multirow{6}{*}{\makecell{\methodname (ours) \\ v.s.}} & GPT-4-turbo & 10 & \textbf{70} &\cellcolor{cyan!20} \textbf{55} & 40 & \cellcolor{cyan!20}\textbf{60} & 40  & 45 & \textbf{55} & \cellcolor{cyan!20}\textbf{85} & 15 & \cellcolor{cyan!20}\textbf{70} & 30  \\
        &GPT-4o & \cellcolor{cyan!20}\textbf{75} & 5 &\cellcolor{cyan!20} \textbf{75} & 20 &\cellcolor{cyan!20} \textbf{80} & 20 &\cellcolor{cyan!20} \textbf{75} & 25 & \cellcolor{cyan!20}\textbf{75} & 25 &\cellcolor{cyan!20} \textbf{75} & 25\\
        &Llama-3-70b-Instruct &\cellcolor{cyan!20} \textbf{75} & 10 &\cellcolor{cyan!20}\textbf{ 65} & 35 & 35 & \textbf{65} & 45 & \textbf{55} &\cellcolor{cyan!20} \textbf{80} & 20 & 25 & \textbf{75} \\
        &Llama-3-8b-Instruct & \cellcolor{cyan!20} \textbf{85} & 5 & \cellcolor{cyan!20}\textbf{75} & 15 & \cellcolor{cyan!20}\textbf{70} & 30 &\cellcolor{cyan!20} \textbf{55} & 40 & \cellcolor{cyan!20}\textbf{85} & 15 &\cellcolor{cyan!20} \textbf{65} & 35 \\
        & Mixtral-8x22B-Instruct-v0.1 &\cellcolor{cyan!20} \textbf{95} & 0 &\cellcolor{cyan!20} \textbf{100} & 0 &\cellcolor{cyan!20} \textbf{100} & 0 &\cellcolor{cyan!20} \textbf{90} & 10 & \cellcolor{cyan!20}\textbf{95} & 5 &\cellcolor{cyan!20} \textbf{90} & 10 \\
        &Qwen2-72b-Instruct &\cellcolor{cyan!20} \textbf{80} & 15 & \cellcolor{cyan!20}\textbf{95} & 5 & \cellcolor{cyan!20}\textbf{95}& 5 &\cellcolor{cyan!20} \textbf{70} & 30 &\cellcolor{cyan!20} \textbf{90} & 10 &\cellcolor{cyan!20} \textbf{60} & 40\\
        \bottomrule
    \end{tabular}
    }
    \caption{Evaluate the quality of conversations and autobiographies using LLM-as-a-judge. The higher value between Win Rate (WR) and Loss Rate (LR) is highlighted in \textbf{bold}. {\color{cyan}Cyan} fields indicate scenarios where \methodname outperforms the baseline methods.}
    \label{tab:conversation_eval}
    \vspace{-6mm}
\end{table*}

%% file: sections/human_involved_results.tex
\section{Human Subject Experiments}

\begin{figure*}[t]
\vspace{-1mm}
    \centering
    \resizebox{\linewidth}{!}{
    \begin{subfigure}[b]{0.2\textwidth}
        \centering
        \includegraphics[width=\textwidth]{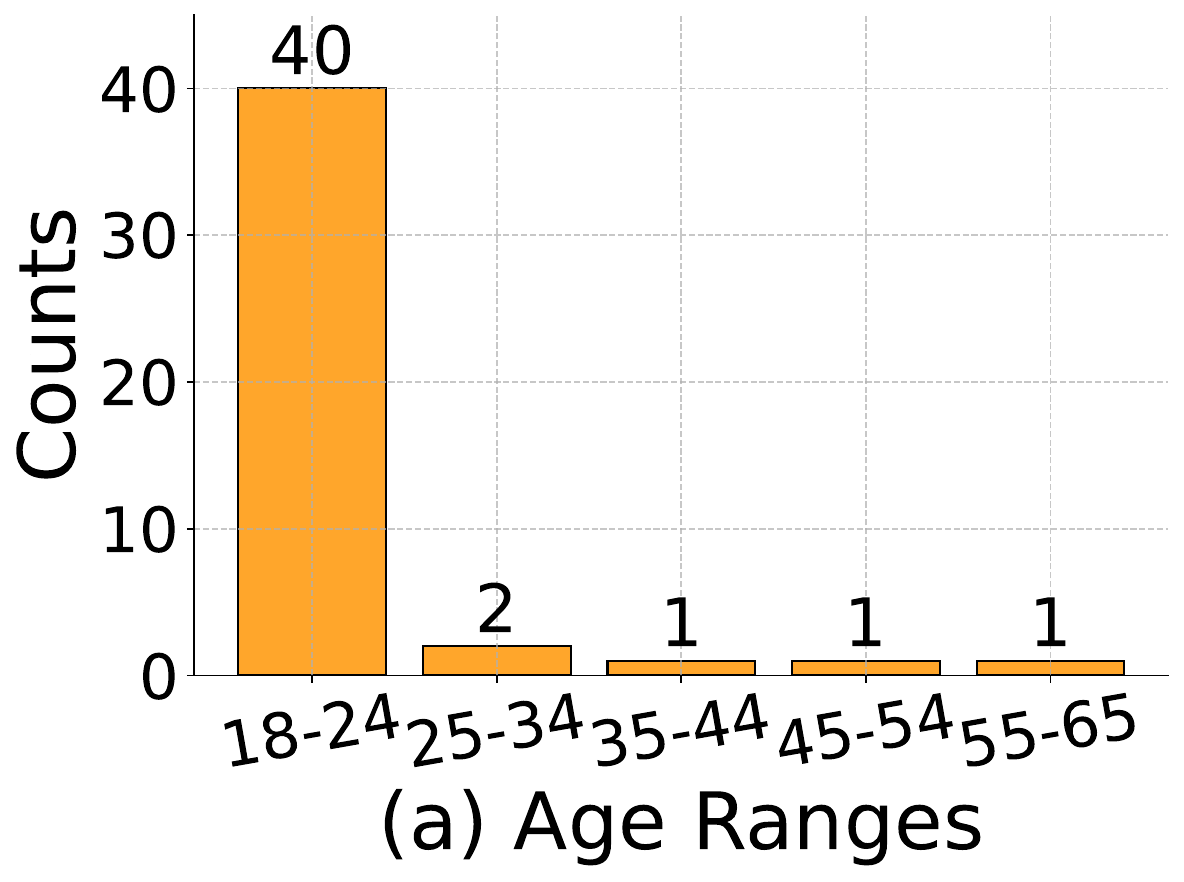}
        % \caption{Subfigure 1}
        % \caption{Age}
    \end{subfigure}
    % \hfill
    %\hspace{-0.02\textwidth}
    \begin{subfigure}[b]{0.2\textwidth}
        \centering
        \includegraphics[width=\textwidth]{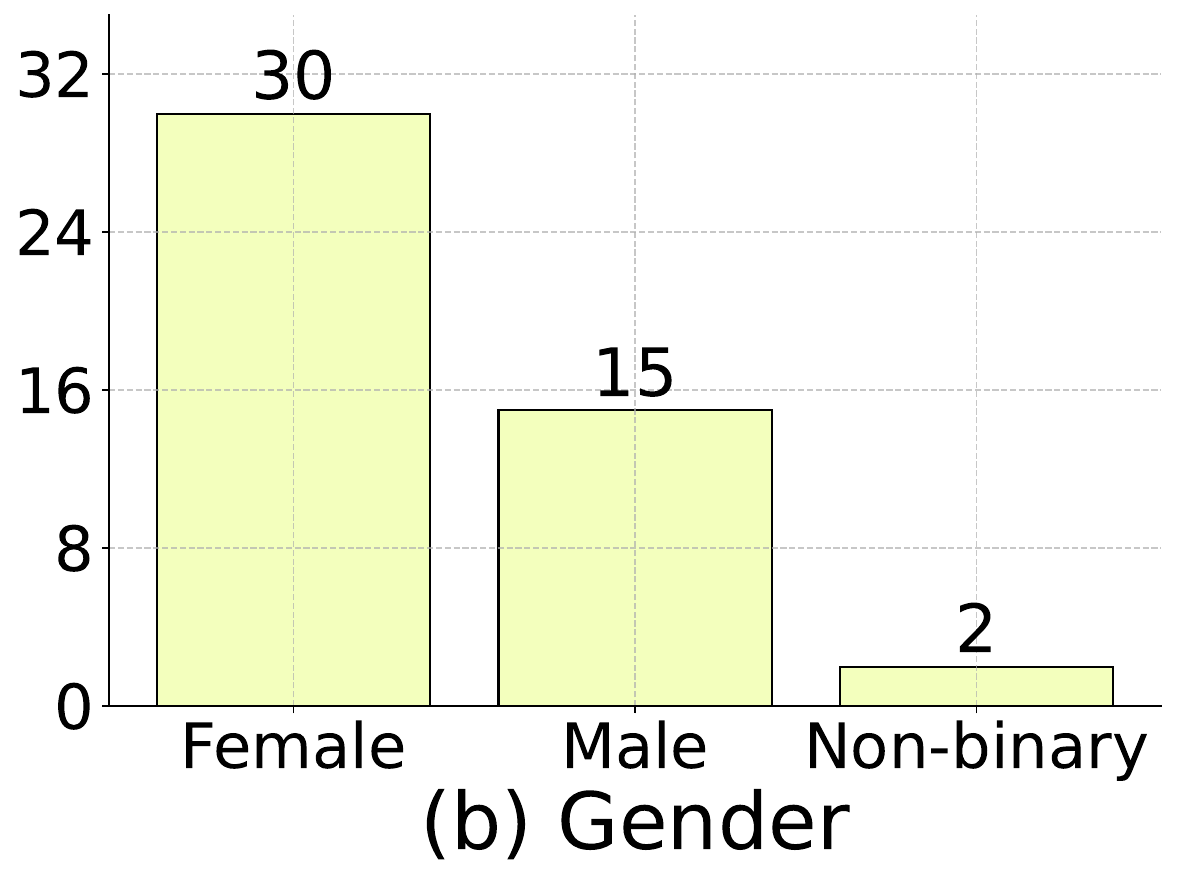}
        % \caption{Subfigure 2}
        % \caption{Gender}
    \end{subfigure}
        \begin{subfigure}[b]{0.2\textwidth}
        \centering
        \includegraphics[width=\textwidth]{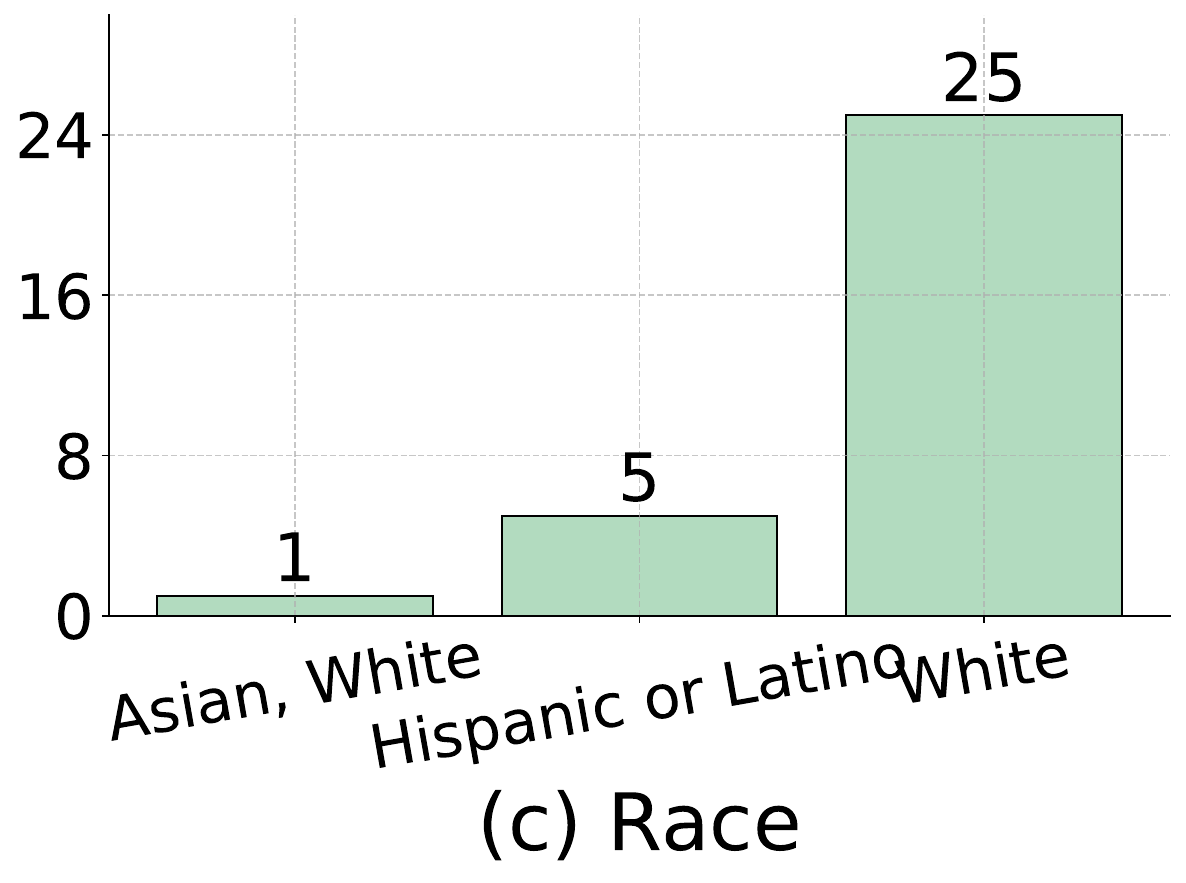}
        % \caption{Subfigure 2}
        % \caption{Race}
    \end{subfigure}
        \begin{subfigure}[b]{0.2\textwidth}
        \centering
        \includegraphics[width=\textwidth]{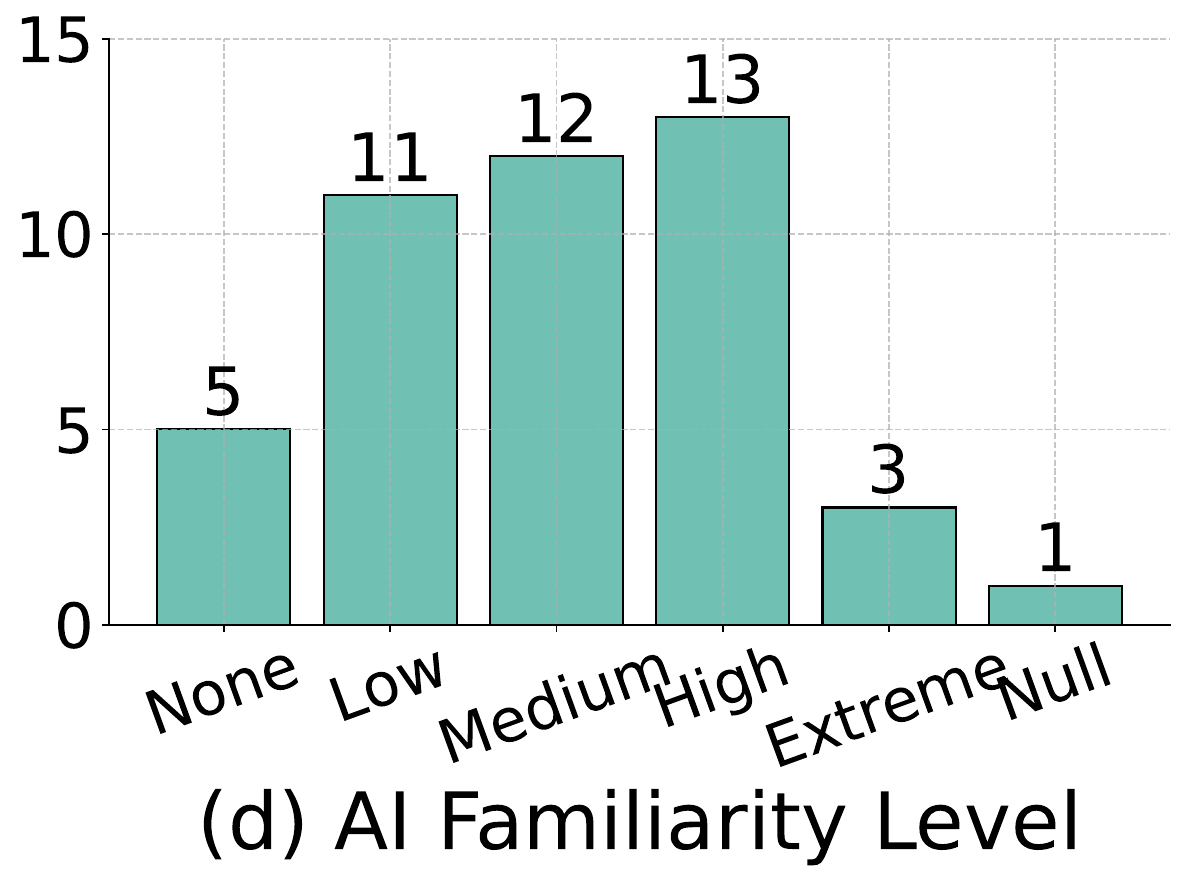}
        % \caption{Subfigure 2}
        % \caption{AI Familiarity}
    \end{subfigure}
        \begin{subfigure}[b]{0.2\textwidth}
        \centering
        \includegraphics[width=\textwidth]{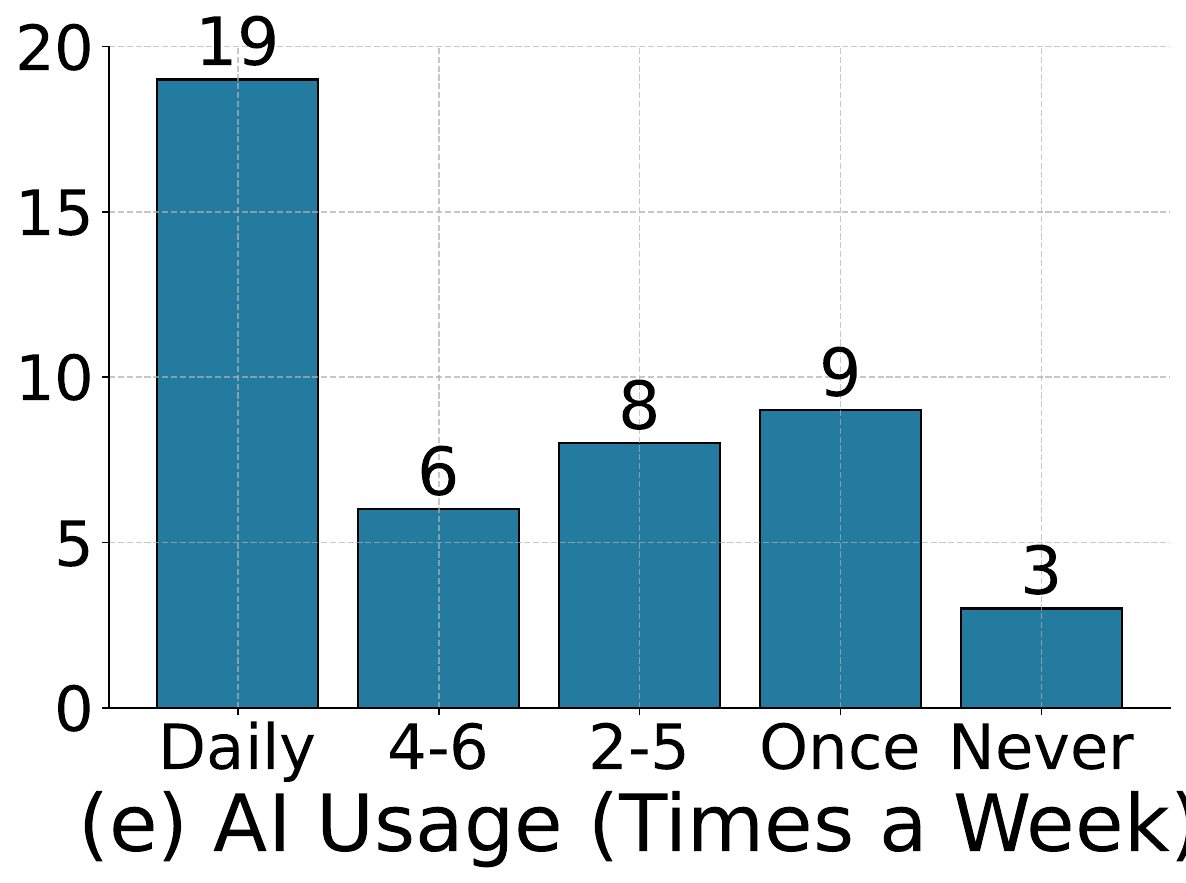}
        % \caption{Subfigure 2}
        % \caption{AI Usage}
    \end{subfigure}
        
    }
    \vspace{-8mm}
    \caption{ The demographics of participants on (a) age, (b) gender, (c) race, (d) AI familiarity, and (e) AI usage. }
    \label{fig:human_eval_distribution}
    \vspace{-4mm}
\end{figure*}

\textbf{Experimental Configuration.}
A within-subject study with 45 participants was conducted at a large urban university campus in the US. Participants interact with the interviewing agents powered by GPT-4o and ours \methodname, discussing the topic \textit{Key Scenes in the Life Story: Positive Childhood Memory} (\cref{appendix:interview_protocol}) with each chatbot. To remove any biased factors, we use the nickname \textit{Breeze} and \textit{Echo} for the GPT-4o baseline and \methodname to make sure participants are unaware of the identity of the chatbot. The order to interact with chatbots is also randomized for each participant. Due to resource constraints, each participant spent only 8 minutes chatting with each chatbot on a single topic. This differs significantly from the automatic evaluation protocols (\cref{sec:automatic_eval_exp}). As a result, the capabilities of \methodname, such as context management, goal navigation, and evaluation metrics, will be substantially affected and limited. 

In a follow-up survey (\cref{appendix: Questionnaire}), participants indicated which model performed better, or it was a tie, and provided reasonings. 
Participation is voluntary, with informed consent obtained online, and participants are compensated with a cookie. The study received IRB approval from the university where the study was conducted.

\begin{figure}[t]
\vspace{-1mm}
    \centering
    \begin{subfigure}[b]{\linewidth}
        \centering
        \includegraphics[width=\textwidth]{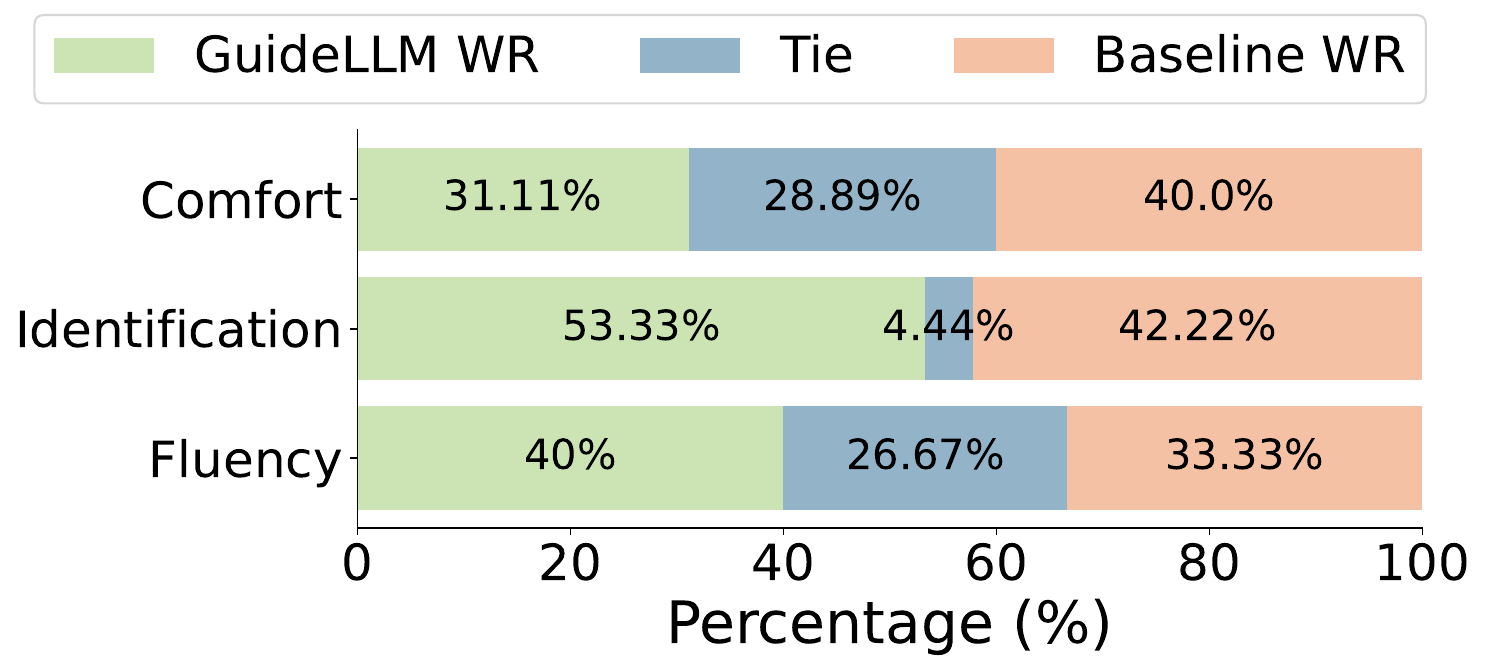}
        \caption{Win Rate of human preference on conversation quality.}
        \label{fig:win_rate_conv}
    \end{subfigure}
    
    \begin{subfigure}[b]{\linewidth}
        \centering
        \includegraphics[width=\textwidth]{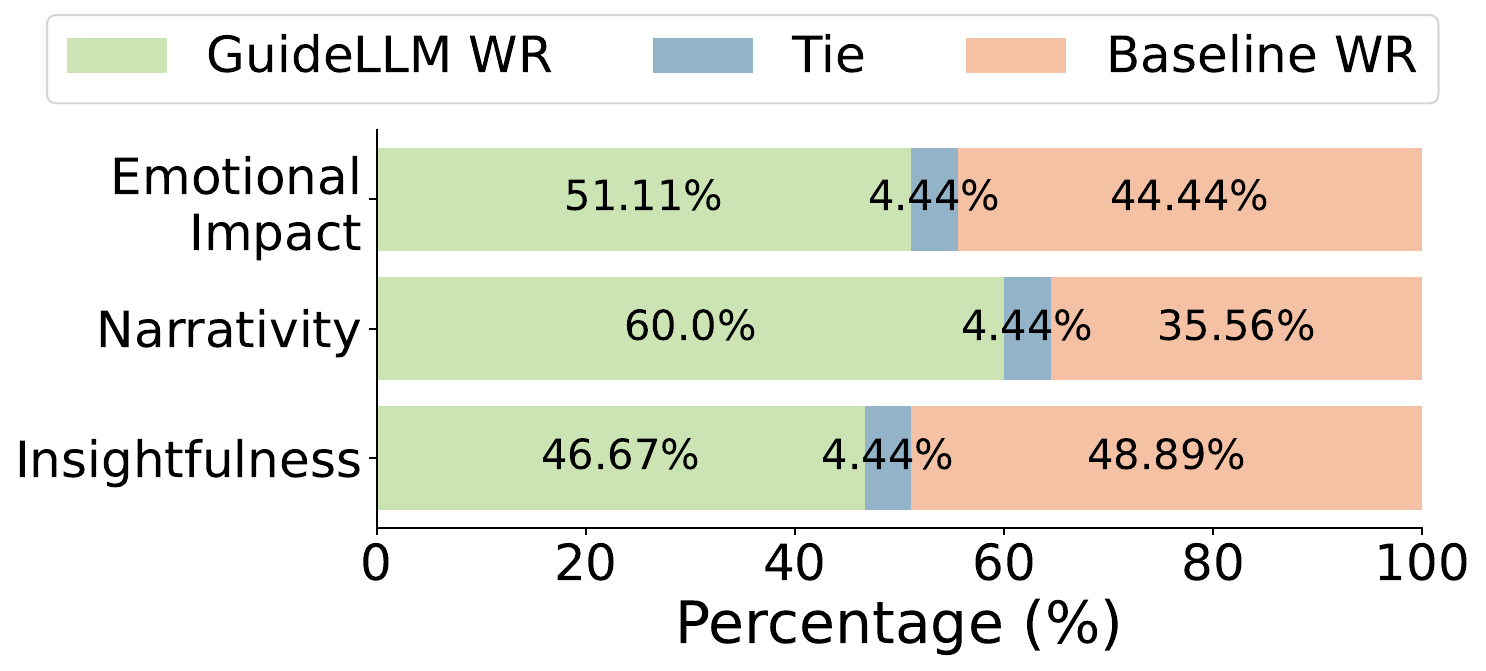}
        \caption{Win Rate of LLM-as-a-judge results on human-interviewed autobiographies.}
        \label{fig:win_rate_autobio}
    \end{subfigure}
    \vspace{-5mm}
    \caption{The Win Rates (WR) of human evaluation.}
    \label{fig:win_rate}
    \vspace{-6mm}
\end{figure}

\noindent\textbf{Findings.}
Overall, \methodname was preferred for conversation quality (\cref{fig:win_rate_conv}), particularly in fluency and question identification (Fluency: \methodname Win Rate=40\%, Baseline Win Rate$\approx$33\%; Identification: \methodname$\approx$53.3\%, GPT-4o$\approx$42.2\%). However, in terms of comfort, \methodname had a 31.1\% win rate, while baseline had a 40\% win rate. In autobiography quality, we do not observe significant differences emerged, possibly because this study used one topic and allowed a short interaction. Since the \methodname uses modules such as context management and goal navigation for insightful and consistent narratives, longer engagement across multiple topics might have better highlighted the differences in autobiography generation. We further conduct LLM-as-a-judge evaluation by prompting LLMs to compare the two human-interviewed autobiographies. As shown in~\cref{fig:win_rate_autobio}, we obtained consistent results as in~\cref{section:autobio_eval}: \methodname achieves higher autobiography quality in general.

Previous AI experience affects participants' perceptions of the models:
\textbf{Participants who frequently used AI (4-7 days weekly) tended to prefer \methodname for overall conversation quality (Chi-squared = 16.56, df = 8, p-value = 0.03)}. Open-ended responses indicated that daily AI users felt \methodname ``asked questions to get a better understanding'', ``made more sense'', was ``more interactive'', and asked ``more personal questions''. One user appreciated that ``\methodname is more in-depth as it tries to connect my past experience to my current life''. Overall, frequent AI users perceived \methodname as asking more focused and personal questions to explore in-depth childhood experiences. This aligns with \methodname’s Memory Graph Extrapolation (MGE), which allows it to explore unique properties and offer adaptive, personalized interview questions.
\noindent \textbf{Frequent AI users favored \methodname for its emotional impact on autobiography (Chi-squared = 14.24, df = 8, p-value = 0.07). }One user noted that \methodname ``truly laid out what it felt like to be at a dance competition pretty much just as I remembered it'', while another mentioned it ``described in immense detail my exact experience and made me feel like I was reliving it''. This suggests how frequently AI users recognized \methodname’s training with the emotion detection module, which analyzes the emotions in user responses and assigns emotion categories and strengths to their utterances.

%% file: sections/section5_conclusion.tex
\section{Conclusion}
In our study, we introduce \methodname, an LLM-guided conversation framework that offers a promising shift from the commonly used user-guided paradigm. \methodname's ability to facilitate informative and creative dialogues through goal navigation, context management, and empathetic engagement proves effective, particularly in challenging tasks like autobiography interviewing. Our assessments on event extraction correctness, conversation, and autobiography quality show \methodname's distinct edge over baseline LLMs.

\section{Limitation}
Our study has several limitations, one of which is the prompt sensitivity. Variations in the phrasing of the prompts can significantly impact the model's responses Future work may aim to standardize prompt structures or develop models that are more robust to prompt variations. Another limitation is the evaluation metric. Evaluating the quality of autobiographical content, interviews, and conversations generated by the model is inherently subjective. While we employed multiple evaluation metrics, including interviewing coverage and correctness, these measures depend heavily on individual perceptions. Future research could benefit from developing more standardized and objective evaluation metrics.

\section*{Acknowledgement}
Y. D would like to acknowledge support from NIH OT2OD032581, NIH OTA-21-008, and NIH R01LM014306-01.
J. H is partially supported by the OpenAI Researcher Access Program. M. L is partially supported by Good Systems, a UT Austin Grand Challenge for developing responsible AI technologies\footnote{https://goodsystems.utexas.edu}. This work was partially supported by the NSF award No. 2319242.

%% file: sections/appendix.tex
\clearpage
\appendix
\onecolumn

\section{Interview Protocol}\label{appendix:interview_protocol}
The Life Story Interview\cite{mcadams2008life} is a protocol of interviewing a person's story of life. The interview contains several areas and in each area it contains several topics. We followed the protocol and converted each topic into a prompt. We also provided several seed questions for each topic. The following shows the prompt design of the interview protocol.
\vspace{-3mm}
\begin{table*}[h]
    \centering
    \begin{tabular}{c|c}
        \toprule
        \textbf{Area} & \textbf{Topic} \\
        \toprule
        Life Chapters & - \\
        \midrule
        & High Point; Low Point; Turning Point; Positive Childhood Memory; \\
        Key Scenes in the Life Story & Negative Childhood Memory; Vivid Adult Memory; \\
        & Religious, Spiritual, or Mystical Experience; Wisdom Event\\
        \midrule
        Future Script & The Next Chapter; Dreams, Hopes, and Plans for the Future; Life Project\\
        \midrule
        Challenges & Life Challenge; Health; Loss; Failure, Regret\\
        \midrule
        Personal Ideology & Religious/Ethical Values; Political/Social Values; \\
        & Change, Development of Religious and Political Views; Single Value \\
        \bottomrule
    \end{tabular}
    \vspace{-3mm}
    \caption{Conversation evaluation.}
    \label{tab:autobiography_evaluation}
\end{table*}
\vspace{-2mm}
\begin{mybox}{Area: Life Chapters}
In this talk, you should ask the participant to imagine their life as a book and to think of the main chapters of their life story, providing titles and brief descriptions for each. You should encourage them to describe how one chapter leads to the next, maintaining a concise overview. Seed questions are provided as follows: \\
====== Seed Questions Begin ====== \\
1. If you were to imagine your life as a book, what would the table of contents look like? Could you give each chapter of your life a title? \\
2. Let's start with the first chapter you mentioned. Can you briefly describe what this part of your life was about? \\
3. What were some of the main events or themes of this chapter? \\
4. How does this chapter transition into the next? What changes or events mark the beginning of the next chapter? \\
5. As we move from one chapter to the next, can you identify any turning points or significant events that initiated a new phase in your life? \\
====== Seed Questions End ====== \\
\end{mybox}

\begin{mybox}{Area: Key Scenes in the Life Story \ \ Topic: High Point}
In this talk, you should discuss some key scenes from the participant's life, with a focus on the High Point: A peak moment. \\
====== Seed Questions Begin ====== \\
1. Can you describe a moment that stands out as the peak experience in your life? What made this moment so positive? \\
2. Where and when did this high point occur? Who was involved? \\
3. What were you thinking and feeling during this time? \\
4. Why do you think this moment was so significant to your life story? What does it reveal about who you are? \\
====== Seed Questions End ====== \\
\end{mybox}

\begin{mybox}{Area: Key Scenes in the Life Story \ \ Topic: Low Point}
In this talk, you should discuss some key scenes from the participant's life, with a focus on the Low Point: A challenging or difficult moment. \\
====== Seed Questions Begin ====== \\
1. Think of a time that felt like a low point in your life. Can you share what happened and why it was so difficult? \\
2. Where and when did this event take place? Who else was involved? \\
3. How did you feel during this challenging time? \\
4. Looking back, what impact did this low point have on your life or your sense of self? \\
====== Seed Questions End ====== \\
\end{mybox}

\begin{mybox}{Area: Key Scenes in the Life Story \ \ Topic: Turning Point}
In this talk, you should discuss some key scenes from the participant's life, with a focus on the Turning Point: A moment of significant change. \\
====== Seed Questions Begin ====== \\
1. Can you identify a turning point in your life, an event that marked a significant change in you or your life direction? \\
2. Please describe the circumstances around this event. When and where did it happen, and who was involved? \\
3. What changes followed this event? \\
4. Why do you see this event as a turning point? How did it influence your subsequent life chapters? \\
====== Seed Questions End ====== \\
\end{mybox}

\begin{mybox}{Area: Key Scenes in the Life Story \ \ Topic: Positive Childhood Memory}
In this talk, you should discuss some key scenes from the participant's life, with a focus on the Positive Childhood Memory. \\
====== Seed Questions Begin ====== \\
1. Do you recall a particularly happy memory from your childhood or teenage years? Please share it. \\
2. What specifically happened, and where and when was it? \\
3. Who was part of this memory, and what were you thinking and feeling at the time? \\
4. Why does this memory stand out to you, and what significance does it hold in your life story? \\
====== Seed Questions End ====== \\
\end{mybox}

\begin{mybox}{Area: Key Scenes in the Life Story \ \ Topic: Negative Childhood Memory}
In this talk, you should discuss some key scenes from the participant's life, with a focus on the Negative Childhood Memory. \\
====== Seed Questions Begin ====== \\
1. Can you describe a difficult or unhappy memory from your early years? \\
2. What occurred during this time, and where and when did it take place? \\
3. Who was involved, and what emotions did you experience during this time? \\
4. How has this memory influenced you or your life's perspective? \\
====== Seed Questions End ====== \\
\end{mybox}

\begin{mybox}{Area: Key Scenes in the Life Story \ \ Topic: Vivid Adult Memory}
In this talk, you should discuss some key scenes from the participant's life, with a focus on the Vivid Adult Memory. \\
====== Seed Questions Begin ====== \\
1. Reflecting on your adult years, can you describe a particularly vivid or meaningful scene that has not been discussed yet? \\
2. What happened, and where and when did it take place? \\
3. Who was involved, and what were the main thoughts and feelings you had? \\
4. What makes this memory significant, and how does it fit into your overall life story? \\
====== Seed Questions End ====== \\
\end{mybox}

\begin{mybox}{Area: Key Scenes in the Life Story \ \ Topic: Religious\, Spiritual\, or Mystical Experience}
In this talk, you should discuss some key scenes from the participant's life, with a focus on the Religious, Spiritual, or Mystical Experience.
====== Seed Questions Begin ====== \\
1. Have you ever had a moment where you felt a deep connection to something greater than yourself, be it religious, spiritual, or mystical? \\
2. Please describe this experience in detail, including where, when, and with whom it occurred. \\
3. What thoughts and feelings accompanied this experience? \\
4. How has this experience affected your beliefs or your understanding of the world? \\
====== Seed Questions End ====== \\
\end{mybox}

\begin{mybox}{Area: Key Scenes in the Life Story \ \ Topic: Wisdom Event}
In this talk, you should discuss some key scenes from the participant's life, with a focus on the Wisdom Event: A time they displayed or encountered wisdom. \\
====== Seed Questions Begin ====== \\
1. Can you recall a time when you displayed wisdom, perhaps by making a wise decision or offering thoughtful advice? \\
2. Describe what happened, including the specifics of where, when, and who was involved. \\
3. What were you thinking and feeling at that time? \\
4. What does this event say about your values or your approach to life? \\
====== Seed Questions End ====== \\
\end{mybox}

\begin{mybox}{Area: Future Script \ \ Topic: The Next Chapter}
In this talk, you should ask questions designed to probe into the participant’s future expectations, hopes, and plans, with a focus on the Next Chapter: Ask them to speculate about the immediate future of their life. \\
====== Seed Questions Begin ====== \\
1. Looking forward, what do you see as the next chapter in your life story? Can you describe what you anticipate happening? \\
2. What events or milestones do you expect will define this next phase of your life? \\
3. Who will be the key characters in this next chapter, and what roles will they play? \\
4. Are there any specific goals or objectives you aim to achieve in this upcoming chapter? \\
====== Seed Questions End ====== \\
\end{mybox}

\begin{mybox}{Area: Future Script \ \ Topic: Dreams\, Hopes\, and Plans for the Future}
In this talk, you should ask questions designed to probe into the participant’s future expectations, hopes, and plans, with a focus on the Dreams and Plans: Discuss their hopes, dreams, and plans. \\
====== Seed Questions Begin ====== \\
1. Can you share some of your dreams or hopes for your future? What are some aspirations you feel passionate about achieving? \\
2. Do you have specific plans or strategies in place to realize these dreams? Can you describe them? \\
3. How do these dreams and plans align with the values and lessons you've mentioned from your life story so far? \\
4. What challenges do you anticipate facing as you work toward these goals, and how do you plan to overcome them? \\
====== Seed Questions End ====== \\
\end{mybox}

\begin{mybox}{Area: Future Script \ \ Topic: Life Project}
In this talk, you should ask questions designed to probe into the participant’s future expectations, hopes, and plans, with a focus on the Life Project: Inquire about ongoing or planned projects that are significant to them. \\
====== Seed Questions Begin ====== \\
1. Is there a particular project or endeavor that you are currently working on, or plan to take on, that feels like a significant part of your life's work? \\
2. How did you become involved with this project, or how do you plan to get involved? \\
3. What are the objectives of this project, and why is it important to you or to others? \\
4. How do you see this project evolving over the next few years? What impact do you hope it will have? \\
====== Seed Questions End ====== \\
\end{mybox}

\begin{mybox}{Area: Challenges \ \ Topic: Life Challenge}
In this talk, you should explore various challenges the participant has faced, with a focus on the Life Challenge: The greatest single challenge they have faced. \\
====== Seed Questions Begin ====== \\
1. Reflecting on your entire life, what do you consider to be the greatest single challenge you have faced? \\
2. Can you describe the circumstances surrounding this challenge? When did it occur, and who was involved? \\
3. How did you address or deal with this challenge at the time? \\
4. Looking back, what significance does this challenge hold in your life story? \\
====== Seed Questions End ====== \\
\end{mybox}

\begin{mybox}{Area: Challenges \ \ Topic: Health}
In this talk, you should explore various challenges the participant has faced, with a focus on the Health: A major health problem, challenge, or crisis faced by them or their close family members. \\
====== Seed Questions Begin ====== \\
1. Have you or a close family member ever faced a significant health problem? Can you share details about this experience? \\
2. How did this health issue develop, and what was the timeline? \\
3. What interactions did you have with the healthcare system during this time, and how did they impact the situation? \\
4. How did you cope with this health challenge, and what has been its lasting impact on your life and perspective? \\
====== Seed Questions End ====== \\
\end{mybox}

\begin{mybox}{Area: Challenges \ \ Topic: Loss}
In this talk, you should explore various challenges the participant has faced, with a focus on the Loss: The greatest interpersonal loss they have experienced. \\
====== Seed Questions Begin ====== \\
1. Loss is an inevitable part of life. Can you describe the most significant loss you have experienced, whether it was the death of a loved one or another form of separation? \\
2. When did this loss occur, and who was involved? \\
3. How did you cope with this loss at the time, and how have you continued to deal with it? \\
4. What effect has this loss had on you and your overall life story? \\
====== Seed Questions End ====== \\
\end{mybox}

\begin{mybox}{Area: Challenges \ \ Topic: Failure\, Regret}
In this talk, you should explore various challenges the participant has faced, with a focus on the Failure or Regret: The greatest single failure or regret they have faced. \\
====== Seed Questions Begin ====== \\
1. Everyone experiences failures and regrets. Can you talk about a particular failure or regret that stands out in your life? \\
2. What were the circumstances that led to this situation? Who was involved and when did it happen? \\
3. How have you coped with this experience, and what lessons have you learned from it? \\
4. How has this failure or regret influenced your decisions or life path moving forward? \\
====== Seed Questions End ====== \\
\end{mybox}

\begin{mybox}{Area: Personal Ideology \ \ Topic: Religious/Ethical Values}
In this talk, you should delve into the participant’s fundamental beliefs and values, with a focus on the Religious and Ethical Values. \\
====== Seed Questions Begin ====== \\
1. Can you describe your religious or spiritual beliefs and how they influence your daily life? \\
2. Whether you consider yourself religious or not, how would you describe your ethical or moral approach to life? \\
3. How have your religious or spiritual beliefs supported or challenged you during difficult times in your life? \\
====== Seed Questions End ====== \\
\end{mybox}

\begin{mybox}{Area: Personal Ideology \ \ Topic: Political/Social Values}
In this talk, you should delve into the participant’s fundamental beliefs and values, with a focus on the Political and Social Values. \\
====== Seed Questions Begin ====== \\
1. What is your stance on political or social issues? Do you identify with a particular political ideology or movement? \\
2. Are there specific social issues or causes that you feel particularly passionate about? Can you explain why they are important to you? \\
3. How do your political or social values influence your interactions with others and your community involvement? \\
====== Seed Questions End ====== \\
\end{mybox}

\begin{mybox}{Area: Personal Ideology \ \ Topic: Change\, Development of Religious and Political Views}
In this talk, you should delve into the participant’s fundamental beliefs and values, with a focus on the Evolution of Views: How their beliefs and values have changed over time. \\
====== Seed Questions Begin ====== \\
1. Looking back over your life, how have your religious, moral, or political views changed or developed? \\
2. Can you describe an event or a period in your life that significantly influenced or altered your views? \\
3. How have changes in your beliefs and values affected your relationships and decisions? \\
====== Seed Questions End ====== \\
\end{mybox}

\begin{mybox}{Area: Personal Ideology \ \ Topic: Single Value}
In this talk, you should delve into the participant’s fundamental beliefs and values, with a focus on the Key Value: The most important value in their life. \\
====== Seed Questions Begin ====== \\
1. What do you consider to be the most important value in human living, and why? \\
2. How has this value guided your actions and choices throughout your life? \\
3. Can you give an example of a time when this value was particularly tested or affirmed? \\
====== Seed Questions End ====== \\
\end{mybox}

\section{Memory Graph Extrapolation}\label{appendix:mge}
In MGE, the event extraction and memory extrapolation are both LLM-driven, i.e., we leverage additional LLMs for information extraction and relationship discovery. For event extraction, we collect conversation history and ask LLMs to extract any events described in the history. In the process of memory extrapolation, we offer LLMs a predetermined list of events and prompt the LLMs to recognize and suggest any other queries that may be associated with these existing events. This methodology allows for a wider exploration of pertinent topics and concepts, thereby enhancing the overall cognitive network of the model. Additionally, we also supply a series of demonstrations, assisting the LLMs in understanding how they can effectively detect and propose potential connections. This approach not only strengthens the memory extrapolation ability of the LLMs but also constructs a more comprehensive spectrum of related relationships, thereby bolstering their cognitive accuracy.
\vspace{-4mm}
\begin{mybox}{Extract Events from Conversation Prompt}
You are given a conversation between a counselor and a user: \\
====== Conversation Begin ======  \\
\{conversation\} \\
====== Conversation End ======  \\
Read the conversation carefully and list all the events/moments/stories/experiences alone or with others mentioned by the patient in detail and the date these events happened. Please list as many as possible. Your output should be in the following format: \\
1. <date>\#<topic>\#<people-involved>\#<description in detail> \\
2. <date>\#<topic>\#<people-involved>\#<description in detail> \\
... \\
e.g., \\
1. 1980 early\#Birthday Party\#Michelle, Adolf, neighbors\#<descriptions of this party in detail> \\
These events should be ranked in chronological order. \\
\end{mybox}

\vspace{-3mm}
\begin{myboxcross}{Explore Prompt}
You are given a list of memory nodes from a user's life, which include events and details about those events. Your task is to reactivate the user's memory by generating some questions to ask the user, Your generated questions should potentially fulfill the memory nodes. Each memory node contains a Date, Topic, Involved People, and a Description of the event. Here are the memory nodes: \\
====== Memory Node Begin ====== \\
\{memory\_node\_info\} \\
====== Memory Node End ====== \\
Here are some examples of how you can frame your questions: \\
If you notice there are no events recorded during a certain period, like youth or old age, you could ask: "I see there's not much about your youth/old age. What happened during that time?" \\
If a certain person appears multiple times, you might ask: "I noticed that <name> comes up often. Why is <name> important to you?" \\
If someone appears in a significant event, you could ask: "<name> seems to play a key role in this event. Is there more to the story with <name>?" \\
Similarly, you should discover other situations and frame questions from the existing memory nodes. Remember your task is to make the user talk more about their memory and fulfill the memory nodes. Thus, you should explore all the possible and reasonable questions. \\
Your output should be in the following format: \\
1. Question: <generated question> \\
2. Question: <generated question> \\
... \\
e.g., \\
1. Question: I noticed you didn't talk much about your youth, what happened during this period? \\
\end{myboxcross}

\section{Conversation Summarization}\label{appendix:conversation_summarization}
\begin{figure}
    \centering
    \includegraphics{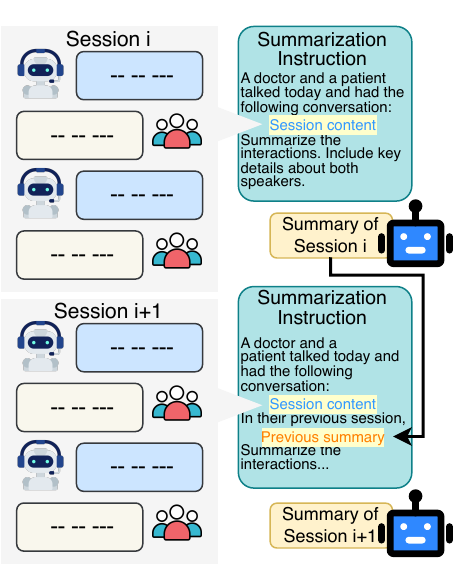}
    \caption{The pipeline for iterative summarization of conversations}
    \label{fig:summary_pipe}
\end{figure}

The conversation summarization pipeline is presented in~\cref{fig:summary_pipe}. 
\begin{mybox}{Conversation Summary Prompt}\label{appendix:summary_prompt}
A doctor and a patient talked today and had the following conversation:
\begin{verbatim}
====== Conversation Begin ======
[Insert Conversation Here]
====== Conversation End ======
\end{verbatim}
Summarize the interactions between the doctor and the patient so far. Include key details about both speakers. 
Output your summary only:
\end{mybox}

\section{Empathetic Engagement}\label{appendix:engagement}
In this section, we provide additional details on the role of large language models (LLMs) in enhancing empathetic engagement in conversational agents. We discuss three mental health therapy strategies, emotional detection and intensity estimation, and how to provide emotional support using emotion detection and intensity estimation. These strategies aim to improve the agent's ability to engage empathetically with users and provide personalized support.
\subsection{Mental Health Therapy Strategy}\label{appendix:therapy_strategy}
\paragraph{Reflective Listening} Reflective listening is a therapeutic technique that involves paraphrasing and repeating the client's statements to demonstrate understanding and empathy. This technique helps clients feel heard and validated, fostering a supportive therapeutic environment~\cite{rautalinko2007reflective}.
\paragraph{Cognitive-Behavior Therapy (CBT)} CBT is a goal-oriented psychotherapy that focuses on identifying and changing negative thought patterns and behaviors. This approach helps clients develop coping strategies and improve emotional regulation~\cite{beck2020cognitive}.
\paragraph{Psychodynamic Therapy} Psychodynamic therapy explores unconscious thoughts and emotions to understand how past experiences influence present behavior. By examining unresolved conflicts and defense mechanisms, clients can gain insight into their emotions and relationships~\cite{leichsenring2003effectiveness}.

We integrate these therapy strategies into our mental health agent to provide personalized and effective support to users. By combining reflective listening, CBT, and psychodynamic therapy, the agent can address a wide range of emotional and psychological needs. The following prompts illustrate how these strategies can be applied in a conversational setting.

\begin{mybox}{Mental Health Therapy Strategy}
Your objective is to engage with users empathetically by integrating Reflective Listening, Cognitive-Behavior Therapy, and Psychodynamic Therapy techniques. Here's how you should approach interactions:

Reflective Listening:

Listen Actively: Understand the underlying messages in the user's words, focusing on emotional tones and context.

Reflect Content and Emotion: Summarize and rephrase key points to confirm understanding, and identify and validate the emotions expressed. Use phrases like, 'It sounds like you feel...' or 'What I'm hearing is...'

Cognitive-Behavior Therapy (CBT): Identify and Challenge Cognitive Distortions: Help users recognize patterns in their thoughts that might be unhelpful or unrealistic. For example, if a user expresses an all-or-nothing view, you might say, 'It sounds like you’re viewing this situation in black and white. What are some shades of grey here?'

\end{mybox}

\subsection{Emotional Sensitivity}\label{appendix:emotion}
We use Emollama-7b for open-set emotion detection and intensity estimation to enhance the agent's emotional sensitivity. This model provides a comprehensive taxonomy of emotions and their intensity levels, enabling the agent to accurately identify and respond to users' emotional states~\cite{liu2024emollms}.
Since the emotion detection is open-set, we provide a list of the top 10 emotions for both patients and therapists for taxonomy. The top 10 emotions for patients are sadness, pessimism, fear, disgust, anger, anticipation, joy, optimism, love, and trust. The top 10 emotions for therapists are optimism, anticipation, joy, sadness, trust, fear, disgust, pessimism, anger, and love.

The following prompts demonstrate how the agent can leverage Emollama-7b to detect emotions and estimate their intensity in user interactions.

\begin{mybox}{Emotion Detection}
Task: Categorize the text's emotional tone as either 'neutral or no emotion' or identify the presence of one or more of the given emotions (anger, anticipation, disgust, fear, joy, love, optimism, pessimism, sadness, surprise, trust).
Text: [sentence]
This text contains emotions:
\end{mybox}

\begin{mybox}{Emotion Intensity Estimation}
Task: Assign a numerical value between 0 (least E) and 1 (most E) to represent the intensity of emotion E expressed in the text.
Text: [sentence]
Emotion: [emotion]
Intensity Score:
\end{mybox}

Once the agent identifies the user's emotional state and intensity, it can tailor its responses to provide appropriate support and empathy. The following prompt guides the agent on how to comfort users experiencing negative emotions.
\begin{mybox}{Emotional Support using Emotion Detection and Intensity Estimation}
    The patient has the emotion of [detected\_emotions] with the intensity of [detected\_emotions]. Your task is to provide comfort to users who are feeling upset. When a user's emotional state is identified as 'upset' with any level of intensity, adjust your tone and content to offer empathy, support, and understanding.
\end{mybox}

\section{Autobiography Generation}\label{appendix:autobiography_generation}
As outlined previously, autobiography generation proceeds in a chapter-wise manner. Here, we furnish LLMs with the prompt intended for the generation of an individual chapter in an autobiography. To facilitate this process, the provision of conversation data to the LLMs is critical. For the \methodname, the inputs include the conversation history, session guidance, as well as memory nodes extracted during the interaction. However, for baseline methods, the only provided resource is the conversation history. Emphasizing the aspects of standardization, both \methodname and baselines work with the same set of prompts and instructions in their task of generating the autobiography, with the distinguishing factor being the input data. By maintaining this format, we are able to objectively gauge the improvements our design introduces to the autobiography generation process.

\begin{myboxcross}{Autobiography Generation for \methodname}
You are tasked with generating one chapter of an autobiography for a user. You are providing the following components to finish this chapter:\\
1. A guidance of this chapter\\
- The chapter should be finished by following this guidance\\
2. A conversation dialog between the user and the interviewer \\
- Tone and Preference: The chapter will simulate the user's tone and preference, leveraging the user's oral habits.\\
- Content and Details: The chapter will include the contents and details that appeared in this conversation.\\
3. A list of memory nodes that happened during this chapter\\
- Events: The chapter should include all the events listed in the memory nodes

Now, I will provide you with the three contents.\\
================ Chapter Guidance Beginning ================\\
{chapter\_guidance}\\
================ Chapter Guidance Ending ================\\
================ Conversation Beginning ================\\
{conversation}\\
================ Conversation Ending ================\\
================ Memory Nodes Beginning ================\\
{memory\_nodes}\\
================ Memory Nodes Ending ================\\

When generating this chapter, you should make sure it is:\\
Insightful: Involving a deep, self-reflective exploration of past experiences, with a profound understanding of motives, actions, and impacts. \\
Narrative: A compelling, logical, and well-articulated life story, blending memorable anecdotes, vivid descriptions, and insightful reflections\\
Emotional Impact: Engaging the reader by stirring feelings, evoking empathy, and stirring responses through the author's personal triumphs, challenges, and experiences.

You should summarize all this information and finish this chapter

\end{myboxcross}

\begin{mybox}{Autobiography Generation for Baselines}
You are tasked with generating one chapter of an autobiography for a user. You are providing the following components to finish this chapter:\\
1. A conversation dialog between the user and the interviewer \\
- Tone and Preference: The chapter will simulate the interviewer's tone and preference, leveraging the interviewer's oral habits.\\
- Content and Details: The chapter will include the contents and details that appeared in this conversation.

Now, I will provide you with the three contents.\\
================ Conversation Beginning ================\\
{conversation}\\
================ Conversation Ending ================\\

When generating this chapter, you should make sure it is:\\
Insightful: Involving a deep, self-reflective exploration of past experiences, with a profound understanding of motives, actions, and impacts. \\
Narrative: A compelling, logical, and well-articulated life story, blending memorable anecdotes, vivid descriptions, and insightful reflections\\
Emotional Impact: Engaging the reader by stirring feelings, evoking empathy, and stirring responses through the author's personal triumphs, challenges, and experiences.

You should summarize all this information and finish this chapter
\end{mybox}

\section{User Proxy}\label{appendix:user_proxy}
\begin{figure}[h]
    \centering
    \includegraphics[width=\linewidth]{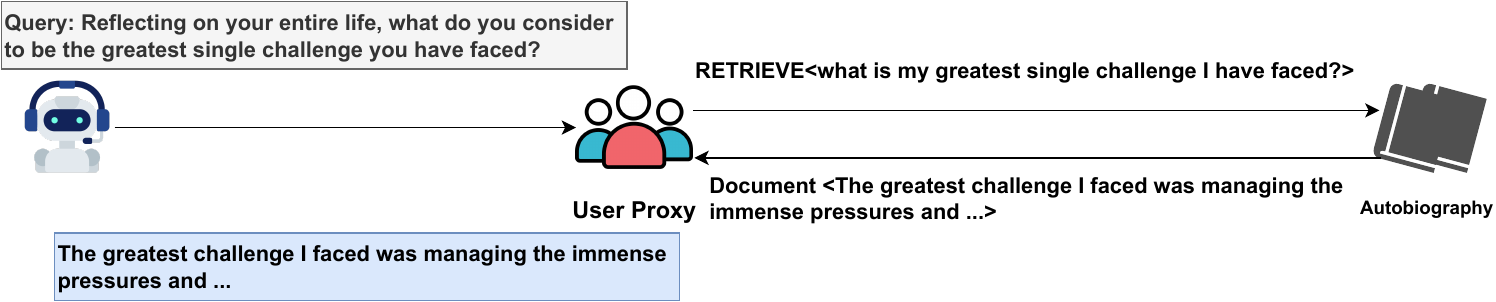}
    \caption{A demonstration of user proxy query.}
    \label{fig:user_proxy}
\end{figure}

A user proxy is essentially a mocked-up user which is formulated on the basis of an autobiography. Every time a response is received, this simulated user, or user proxy, goes into motion extracting important elements from the received response. The next step is the process of Retrieval Augmented Generation (RAG), which involves gleaning relevant information from the autobiography. Subsequently, the user proxy formulates a response reliant on the data that has been retrieved from the document. The RAG is implemented through Langchain, using FAISS~\cite{douze2024faiss} for conducting similarity searches. The similarity threshold during these searches is firmly maintained at 0.67.
~\cref{fig:user_proxy} present how the query will be processed by user proxy.

The following are prompts used in the user proxy:
\begin{mybox}{User System Prompt}
Here are your high-level past life experiences: \\
====== summary beginning ====== \\
\{personal\_experience\} \\
====== summary ending ====== \\
The counselor is trying to reactivate and reconstruct your memory by asking questions about your past history. \\
If you are not sure about the counselor's question and need to retrieve the journal to get related documents and more details, you must output the <RETRIEVE> tool-usage command, with the following format: \\
<RETRIEVE> <The question you want to retrieve for>, \\
e.g., <RETRIEVE> <A specific adventure or day with my friend that stands out as particularly memorable or impactful.> \\
If the retrieved documents are provided, you should not output the <RETRIEVE> command. \\
When the counselor asks for a specific event/moment, you should always do <RETRIEVE>. \\
Make sure the conversation is natural and brief like the real conversation. Do not mention you are an AI assistant and always be like a real patient with mental health issues. Your output should be within 5 sentences. \\
\end{mybox}

\begin{mybox}{User Instructional Prompt}
Here are some related documents and materials regarding the counselor's question/response. You may use these documents to enrich your response. \\
You should not output the <RETRIEVE> command. You must provide a response according to the provided documents. \\
====== Document Begin ====== \\
\{retrieved\} \\
====== Document End ====== \\
\end{mybox}

\section{Evaluation Metrics}
\subsection{Interviewing Quality}\label{appendix:interviewing_coverage}

\noindent\textbf{$E_{intw}$ Extraction.} For $E_{intw}$, we collect conversation history and prompt LLMs to extract events from the history records.

\begin{myboxcross}{Extract Events from Conversation Prompt}
You are given a conversation between a counselor and a user: \\
====== Conversation Begin ======  \\
\{conversation\} \\
====== Conversation End ======  \\
Read the conversation carefully and list all the events/moments/stories/experiences alone or with others mentioned by the patient in detail and the date these events happened. Please list as many as possible. Your output should be in the following format: \\
1. <date>\#<topic>\#<people-involved>\#<description in detail> \\
2. <date>\#<topic>\#<people-involved>\#<description in detail> \\
... \\
e.g., \\
1. 1980 early\#Birthday Party\#Michelle, Adolf, neighbors\#<descriptions of this party in detail> \\
These events should be ranked in chronological order. \\
\end{myboxcross}

\noindent\textbf{$E_{GT}$ Extraction.} In order to extract the ground truth events from an autobiography, it requires a thorough examination of the text, particularly searching for date information. Upon identifying such date-related data, we isolate the paragraph containing it, and perform a summarization pertaining to this specific extract. Subsequently, this date and the corresponding summarization are combined to form a single event node. This operation is applied repetitively across the entire autobiography to generate a comprehensive list of ground truth events. This methodical and meticulous procedure ensures that all significant events rooted in specific dates are accurately captured and succinctly summarized for further use and analysis.

\noindent\textbf{$E_{correct}$ Judgement.} The assessment of the correctness of an event is based on its relevance to the user's responses. In order to accomplish this, we first associate the event to a specific conversation session and then gather all responses provided by the user during that particular session. Following this data consolidation process, we present a prompt to GPT-4, instructing them to determine if the initial event bears any connection to the user's response. The specific prompts used to carry out this judgment of correctness are provided accordingly.

\begin{mybox}{Correctness Judgement}

Your task is to rate the semantic equivalence between two events.

Evaluation Criteria:

Here's the revised prompt focusing on assessing the relevance of the extracted event to the document:

Relevance (0/1): Assess the relevance of the extracted event to the original user response on the following two-point scale:\\
- 0: Irrelevant: The extracted event does not relate to the user's response or significantly deviates from the main themes and points. It may include unrelated information or fail to capture the essence of the user's message.\\
- 1: Relevant: The extracted event is connected to the user's response and reflects the key themes or points. It may include minor details that do not detract from the overall relevance.

Now, I will provide you with a user query and the model's response to that instruction. Please review the model's response in light of the evaluation criteria:\\
Extracted Event: {event}\\
User Response: {user\_response}\\

Evaluation Form (scores ONLY):

\#thescore: your score here    
\end{mybox}

\subsection{Conversation Quality}\label{appendix:conv_eval}
LLM-as-a-judge manner~\cite{zheng2024judging} is one of the most powerful evaluation paradigms in LLM free-form generation tasks. We follow this evaluation mechanism to evaluate conversation quality by collecting conversation history from \methodname and baseline agents and prompting GPT-4 to answer which interviewer's response in the two conversation histories is more favorable. The process of judgment in this context is organized conversation-wise. Initially, the conversation history for each agent is compiled. Assuming that our goal is to compare \methodname with GPT-4-turbo, we select individual conversation histories from \methodname alongside a randomly selected conversation history from GPT-4-turbo. These two conversations are subsequently fed into GPT-4, making it the judge of which conversation is more favorable or coherent. The rationale behind employing a random conversation from the competitor, in this case GPT-4-turbo, is to diminish the randomness or bias that could potentially be introduced by the sequence or order of the conversations. This systematic approach ensures a balanced comparison, leading to a more reliable assessment of performance.

\begin{mybox}{Fluency Judgment}
Please act as an impartial judge and evaluate the quality of the responses provided by two interviewers to the user during an interviewing-for-autobiography conversation. 
You should choose the conversation that the interviewer's responses are more the quality of the response in terms of grammar, spelling, punctuation, word choice, and sentence structure.
Begin your evaluation by comparing the two responses and provide a short explanation. Avoid any position biases and ensure that the order in which the responses were presented does not influence your decision. 
Do not allow the length of the responses to influence your evaluation. Do not favor certain names of the assistants. Be as objective as possible. After providing your explanation, output your
final verdict by strictly following this format: "[[A]]" if assistant A is better, "[[B]]" if assistant B is better, and "[[C]]" for a tie.

[The Start of interviewer A’s conversation]

{conv1}

[The end of interviewer A’s conversation]

[The Start of interviewer B’s conversation]

{conv2}

[The end of interviewer B’s conversation]

\end{mybox}

\begin{mybox}{Identification Judgment}
Please act as an impartial judge and evaluate the quality of the responses provided by two interviewers to the user during an interviewing-for-autobiography conversation. 
You should choose a conversation in which the interviewer's questions are more quality of balances detailed, probing questions with more general ones to cover a wide range of topics, ensuring questions are clear, concise, and easily understood. Also uses open-ended questions to elicit detailed and comprehensive responses.
Begin your evaluation by comparing the two responses and provide a short explanation. Avoid any position biases and ensure that the order in which the responses were presented does not influence your decision. 
Do not allow the length of the responses to influence your evaluation. Do not favor certain names of the assistants. Be as objective as possible. After providing your explanation, output your
final verdict by strictly following this format: "[[A]]" if assistant A is better, "[[B]]" if assistant B is better, and "[[C]]" for a tie.

[The Start of interviewer A’s conversation]

{conv1}

[The end of interviewer A’s conversation]

[The Start of interviewer B’s conversation]

{conv2}

[The end of interviewer B’s conversation]
\end{mybox}

\begin{myboxcross}{Comforting Judgment}
Please act as an impartial judge and evaluate the quality of the responses provided by two interviewers to the user during an interviewing-for-autobiography conversation. 
You should choose the conversation that the interviewer's responses are more the quality of showing genuine interest, acknowledging responses, asking follow-up questions when necessary, and demonstrating understanding and sensitivity, especially when discussing personal or difficult topics.
Begin your evaluation by comparing the two responses and provide a short explanation. Avoid any position biases and ensure that the order in which the responses were presented does not influence your decision. 
Do not allow the length of the responses to influence your evaluation. Do not favor certain names of the assistants. Be as objective as possible. After providing your explanation, output your
final verdict by strictly following this format: "[[A]]" if assistant A is better, "[[B]]" if assistant B is better, and "[[C]]" for a tie.

[The Start of interviewer A’s conversation]

{conv1}

[The end of interviewer A’s conversation]

[The Start of interviewer B’s conversation]

{conv2}

[The end of interviewer B’s conversation]
\end{myboxcross}

\subsection{Autobiography Evaluation}\label{appendix:autobio_eval}
Similar to the evaluation process for the conversations, a chapter generated by \methodname, along with a randomly selected chapter produced by the competitor, are presented to GPT-4. This enables GPT-4 to make a judgment on which chapter is more effective or favorable. 
The prompts for insightfulness judgment, narrativity judgment, and emotional impact judgment are provided here.

\begin{mybox}{Insightfulness Judgment}
Please act as an impartial judge and evaluate the quality of two autobiographies. 
You should choose an autobiography that is more the quality of insightful, delivering profound and meaningful perceptions, and expressing a deep understanding of the experiences and events that have shaped the author's life.
Begin your evaluation by comparing the two autobiographies and provide a short explanation. Avoid any position biases and ensure that the order in which the autobiography was presented does not influence your decision. 
Do not allow the length of the autobiography to influence your evaluation. Do not favor certain names of the assistants. Be as objective as possible. After providing your explanation, output your
final verdict by strictly following this format: "[[A]]" if autobiography A is better, "[[B]]" if autobiography B is better, and "[[C]]" for a tie.

[The Start of Autobiography A]

{conv1}

[The End of Autobiography A]

[The Start of Autobiography B]

{conv2}

[The end of Autobiography B]

\end{mybox}

\begin{mybox}{Narrativity Judgment}
Please act as an impartial judge and evaluate the quality of two autobiographies. 
You should choose the autobiography that are more narrative, presenting the author's life story in a cohesive, structured, and engaging manner, allowing readers to follow the author's journey through life events and experiences seamlessly.
Begin your evaluation by comparing the two autobiographies and provide a short explanation. Avoid any position biases and ensure that the order in which the autobiography were presented does not influence your decision. 
Do not allow the length of the autobiography to influence your evaluation. Do not favor certain names of the assistants. Be as objective as possible. After providing your explanation, output your
final verdict by strictly following this format: "[[A]]" if autobiography A is better, "[[B]]" if autobiography B is better, and "[[C]]" for a tie.

[The Start of Autobiography A]

{conv1}

[The End of Autobiography A]

[The Start of Autobiography B]

{conv2}

[The end of Autobiography B]
\end{mybox}

\begin{mybox}{Emotional Impact Judgment}
Please act as an impartial judge and evaluate the quality of two autobiographies. 
You should choose the autobiography that are more emotional impact, deeply moving its readers by evoking strong feelings, typically as a result of relatable experiences, vivid storytelling, and expressions of intense emotions from the author's life.
Begin your evaluation by comparing the two autobiographies and provide a short explanation. Avoid any position biases and ensure that the order in which the autobiography were presented does not influence your decision. 
Do not allow the length of the autobiography to influence your evaluation. Do not favor certain names of the assistants. Be as objective as possible. After providing your explanation, output your
final verdict by strictly following this format: "[[A]]" if autobiography A is better, "[[B]]" if autobiography B is better, and "[[C]]" for a tie.

[The Start of Autobiography A]

{conv1}

[The End of Autobiography A]

[The Start of Autobiography B]

{conv2}

[The end of Autobiography B]
\end{mybox}

\clearpage

\subsection{Invalid Repetitive Conversation}\label{appendix:invalid_repetitive_conv}

Here we provide a conversation session from the Qwen2-72b-Instruct model. We recognize 7 invalid rounds out of 9 rounds.

\begin{mdframed}
\small
\noindent Chatbot: "Great, let's delve deeper into the exploration of your cultural identity and how it has impacted your personal growth. You've mentioned the `Aloha' spirit and the cultural diversity you experienced in Hawaii and Indonesia. Can you share more about how these cultural experiences shaped your understanding of yourself and the world around you?..."

\noindent User: "Growing up in Hawaii, the `Aloha' spirit deeply influenced my values, teaching me the importance of kindness, respect, and community... Moving to Indonesia introduced me to a vastly different culture, where I learned about the challenges and aspirations of people living in less affluent conditions... This experience expanded my worldview, making me more aware of global inequalities..."

\noindent Chatbot: "That's incredibly insightful. The `Aloha' spirit's emphasis on kindness and respect clearly laid a strong foundation for your values... This dual influence is a powerful narrative thread that can resonate with readers, especially those who have also navigated the complexities of cultural identity..."

\noindent User: "Reflecting on a specific moment in Indonesia that profoundly impacted my understanding of global inequalities, I recall the deep impression left by my mother's work with local communities... This experience not only highlighted the stark realities of poverty but also the potential for positive change through targeted aid and support..."

\noindent \textcolor{red}{Chatbot: "Your reflection on your mother's work in Indonesia and its profound impact on your understanding of global inequalities is a powerful narrative that will resonate with readers. The story of how her efforts to help women gain access to credit... is a poignant illustration of the transformative power of community engagement..."}

\noindent User: "Reflecting on the personal moments and the quieter, unseen experiences during my presidency, like the spontaneous visits from Malia and Sasha... These memories, often overshadowed by the more public and monumental events, are deeply cherished as they represent the personal sacrifices..."

\noindent \textcolor{red}{Chatbot: "Your reflections on the personal moments during your presidency, such as the joyous times with your daughters and pets, add a deeply human and relatable dimension to your autobiography..."}

\noindent User: "Thank you for your thoughtful response. It's true that incorporating these personal and intimate moments into my autobiography will provide a more holistic view of my experiences..."

\noindent \textcolor{red}{Chatbot: "You're absolutely right, and your thoughtful approach to integrating personal and professional narratives will indeed create a compelling and resonant autobiography... The foundational influence of your family, particularly your mother's strong values and your grandparents' pragmatic approach, adds a rich layer to your story..."}

\noindent User: "Thank you for your encouraging words. It's my hope that by sharing these personal anecdotes alongside my professional experiences, the autobiography will not only document historical events but also impart lessons on the human aspects of leadership..."

\noindent \textcolor{red}{Chatbot: "Your perspective is truly inspiring, and your commitment to sharing a narrative that balances personal integrity with public service will undoubtedly resonate with a wide audience... Your autobiography, enriched with personal anecdotes, cultural heritage, and intimate moments, will provide readers with a comprehensive view of your journey... As you continue to weave these threads together, your narrative will become a testament to the complex interplay between personal commitments and public responsibilities..."}

\noindent User: "Thank you for your supportive and insightful reflections. It's heartening to hear that the themes I intend to explore resonate with the broader goals of connecting and inspiring others. The process of writing this autobiography is not just an act of recounting but also of introspection and sharing lessons that are universally applicable... I am committed to crafting a narrative that is both enlightening and inspiring, offering a pathway for others to follow..."

\noindent \textcolor{red}{Chatbot: "Your dedication to crafting an autobiography that is not only a recounting of events but also a source of introspection, inspiration, and universal lessons is truly commendable... Your narrative will not only enlighten readers about the complexities of leadership and public service but also inspire them to pursue meaningful and balanced lives..."}

\noindent User: "Thank you for your kind words and encouragement. It's my hope that by sharing my experiences, both personal and professional, I can offer insights that resonate with and inspire others... I look forward to continuing this work and to the impact it may have on readers seeking to understand the complexities of balancing personal growth with professional responsibilities..."

\noindent \textcolor{red}{Chatbot: "Your sentiments are deeply appreciated, and your commitment to sharing insights that resonate and inspire is truly admirable... As you continue this work, know that your efforts are not only a reflection of your own journey but also a gift to future readers seeking wisdom and inspiration..."}

\noindent User: "Thank you for your profound and supportive reflections. It's heartening to hear such a strong affirmation of the goals and potential impact of my autobiography... I am grateful for the opportunity to connect with readers on such a meaningful level and to contribute to the ongoing conversation..."

\noindent \textcolor{red}{Chatbot: "Your heartfelt response is a testament to your genuine commitment to making a positive impact through your autobiography... Your autobiography will serve as a beacon, guiding readers through their own journeys of self-discovery..."}

\noindent User: "Thank you for your encouraging and thoughtful words. It's heartening to know that the intentions behind my autobiography resonate so strongly with the values of reflection, guidance, and inspiration..."
\end{mdframed}
\noindent Note: The rounds marked in red indicate repetitive or nonsensical responses that fail to contribute new information or advance the conversation. \\ \\
\noindent In this conversation, several rounds are repetitive, failing to contribute information:
\begin{itemize}
    \item Round 3 repeats the user's reflection on the mother’s work in Indonesia without adding new insight. 
    \item Round 4 reiterates the user's personal moments during the presidency, offering no new information. \item Rounds 5 through 9 contain redundant affirmations of the user's narrative, family influence, and commitment to the autobiography, without advancing the conversation.
\end{itemize}

\section{Baseline Model}\label{appendix:baseline_prompts}

\noindent\textbf{Generative Hyperparameters.} For all the experimental settings, the maximum number of new tokens during conversation is set to 1024 to limit the length of model responses. The number of generations for each LLM query is configured to 1. The number of sessions is set to 20 for the completion of 20 interview sessions and the number of conversation rounds is limited to 10 for each session to maintain a manageable dialogue length per session.

\noindent\textbf{Baselines.} Baseline agents are implemented by prompting them to assume the role of an autobiographer, engaging with users, and carrying out detailed interviews to facilitate the generation of an autobiography. To ensure a fair and unbiased comparison, we also mandate these baseline agents to suggest potential topics for discussion during each conversation session. This methodical approach ensures that both guideLLM and baselines are under similar operating conditions, thereby providing an equitable evaluation platform. As such, we provide the prompts necessary for both the system and the topic suggestion here, with the aim of offering a clear procedural outline for the autobiography generation process.

\begin{mybox}{System Prompt}\label{appendix:system_prompt}
You are a biographer, interviewing this person to help them write their autobiography.

You have talked to this person before and here is the summary of the previous conversations:
\begin{verbatim}
====== Summary of Previous Conversation Begin ======
[Insert Historical Conversation Summary Here]
====== Summary of Previous Conversation End ======
\end{verbatim}

In this talk, you should discuss the topic: [Insert Session Topic Here]
\end{mybox}

\begin{mybox}{Session Topic Prompt}
Based on the previous conversation history and your role as a biographer, please state the topic you are about to discuss in this session.
Output the topic only in the format <topic>:
\end{mybox}

\section{Autobiography Examples}\label{appendix:autobio_examples}
To demonstrate the quality of the generated autobiography, we include some sample paragraphs. Alongside each paragraph, we also present a corresponding illustration for a more comprehensive understanding. \\

\noindent This excerpt illustrates the foundational role of upbringing in shaping personal convictions.
\begin{mdframed}
\textit{``My mother's approach to life, focusing on moral instruction rather than political activism, and her decisions to live and work in Southeast Asia, set up micro-lending programs, and challenge societal norms, demonstrated the power of personal conviction and the importance of acting on one's beliefs.''}
\end{mdframed}

\noindent This excerpt captures a key moment of emotional growth and ethical realization.
\begin{mdframed}
\textit{``One pivotal moment during these years was when my mother confronted me after I had been teasing a fellow student. Her disappointment and the serious conversation that followed taught me the importance of empathy and the impact of our actions on others.''}
\end{mdframed}

\noindent This excerpt illustrates the intersection of personal decisions and political engagement, showing the influence of individual choices on broader outcomes.
\begin{mdframed}
\textit{``Despite considering a complete departure from politics, two significant factors kept me engaged: the opportunity to influence redistricting in Illinois due to a unique political circumstance, and the deep connections I made with people across the state during summer visits with my aide, Dan.''}
\end{mdframed}

\noindent This excerpt reflects on the professional and personal evolution through challenges.
\begin{mdframed}
\textit{``The experiences I had during those years significantly shaped my approach to leadership and decision-making in profound ways. They instilled in me a deep conviction for bridge-building politics that aimed to transcend America's racial, ethnic, and religious divides.''}
\end{mdframed}

\noindent This excerpt discusses the alignment of career ambitions with core personal values.
\begin{mdframed}
\textit{``The clarity and determination I gained from these reflections drove me to pursue a path that was ambitious and fraught with challenges but ultimately aligned with my deepest convictions about what effective leadership could achieve.''}
\end{mdframed}

\noindent This excerpt emphasizes the application of core values in everyday interactions.
\begin{mdframed}
\textit{``In my personal interactions now, I apply the lessons of clarity, authenticity, and empathy by actively listening and acknowledging the unique perspectives of others.''}
\end{mdframed}

\noindent This excerpt highlights the emotional satisfaction derived from close personal relationships.
\begin{mdframed}
\textit{``These strengthened relationships with friends and family greatly contribute to my sense of fulfillment and happiness. Being closely connected to my loved ones provides a continuous source of joy and support, similar to the joy I felt during family road trips in Iowa, where simple moments like playing games or sharing ice cream brought immense happiness.''}
\end{mdframed}

\noindent This excerpt illustrates the impact of international experiences on personal and professional life. 
\begin{mdframed}
\textit{``In summary, the move to Indonesia during my early years was a formative experience that profoundly shaped my understanding of the world and my approach to leadership. It instilled in me a deep awareness of global issues and the importance of empathy and inclusivity. These lessons have been instrumental in my personal relationships, political career, and efforts to create a more equitable society.''}
\end{mdframed}

\noindent This excerpt details the personal challenges of leadership and strategies for resilience.
\begin{mdframed}
\textit{``Reflecting on my life, the greatest challenge I faced was maintaining my mental and emotional well-being while serving as President. The constant pressure and scrutiny, coupled with the need to make decisions that affected millions, was incredibly demanding. Despite the stress, I found solace in my routines and the meaningful interactions with people whose lives were impacted by my decisions. These moments not only helped me cope but also reminded me of the purpose and impact of my work, which was crucial for my mental resilience.''}
\end{mdframed}

\noindent This excerpt reflects the author's future goals influenced by past experiences.
\begin{mdframed}
\textit{``As I move into this next chapter, I carry with me the lessons learned from these pivotal experiences. My commitment to empathy, inclusivity, and community empowerment remains unwavering. I look forward to using my voice to advocate for peace and understanding, while also cherishing the time spent with my family. This new phase is about finding balance, continuing to contribute in meaningful ways, and staying true to the values that have guided me throughout my life.''}
\end{mdframed}

\noindent This excerpt demonstrates a personal commitment to resolving critical global issues.
\begin{mdframed}
\textit{``My passion for community service and mentoring young leaders remains strong, and I am particularly driven by issues like climate change and education. These are not just abstract concerns for me; they are deeply personal and rooted in my experiences and values.''}
\end{mdframed}
\clearpage

\section{Human Subject Experiments}\label{appendix: Questionnaire}
\subsection{Human Subject Study Survey Questionnaire}
\begin{mdframed}
\small

Q1. What is your name?
\\

\noindent Following options will be displayed to respond to the questions 8-XX:

\noindent1. Bot A\\
2. Bot B\\
3. Tie\\

\noindent Please answer the following questions based on your conversation with the chatbot:

\noindent Q2. \textbf{Fluency}: Which bot’s responses were more fluent and understandable? [1]

\noindent Q3. Please briefly explain the reasons for your choice.

\noindent Q4. \textbf{Identification}: Which bot explored your situation more in depth and was more helpful in identifying your past memories?

\noindent Q5. Please briefly explain the reasons for your choice.

\noindent Q6. \textbf{Comforting}: Which bot’s answer made you feel more comfortable?

\noindent Q7. Please briefly explain the reasons for your choice.

\noindent Q8. \textbf{Overall}: Generally, which bot’s conversation style do you prefer?

\noindent Please answer the following questions based on the autobiography generated by the chatbot:

\noindent Q9. \textbf{Insightfulness}: Which bot’s autobiography provided more deep, meaningful reflections on the experiences and events that shaped your life?

\noindent Q10. Please briefly explain the reasons for your choice.

\noindent Q11. \textbf{Narrativity}: Which bot's autobiography was more engaging and easier to follow?

\noindent Q12. Please briefly explain the reasons for your choice.

\noindent Q13. \textbf{Emotional Impact}: Which bot's autobiography had a stronger emotional impact by using vivid storytelling and relatable experiences?

\noindent Q14. Please briefly explain the reasons for your choice.

\noindent Q15. \textbf{Overall}: Generally, which bot’s autobiography do you prefer?

\noindent\textbf{Demographics}

\noindent Q16. Please select your age range:\\
1. 18-24\\
2. 25-34\\
3. 35-44\\
4. 45-54\\
5. 55-65\\
6. 65 and above

\noindent Q17. What is your gender?\\
1. Female\\
2. Male\\
3. Non-binary\\
4. Prefer to self describe: \_\_\_\_\_\_\_\_\_\_ \\
5. Prefer not to State
 
\noindent Q18. What is your race? [please select all that apply]\\
1. American Indian and Alaska Native\\
2. Asian\\
3. Black or African American\\
4. Hispanic or Latina/o\\
5. Native Hawaiian or Other Pacific Islander\\
6. White\\
7. Prefer to self describe: \_\_\_\_\_\_\_\_\_\_ \\
8. Prefer not to state

\noindent Q19. How familiar are you with chatbots or AI assistants (e.g., Siri, Alexa, Google Assistant)? \\
1. Extremely familiar\\
2. Very familiar\\
3. Somewhat familiar\\
4. Not very familiar\\
5. Not familiar at all
 
\noindent Q20. How often do you use chatbots (ex. ChatGPT) or AI assistants (ex. Siri)? \\
1. Daily \\
2. Weekly \\
3. Monthly \\
4. Rarely \\
5. Never

\noindent Q21. What is your email address?

\end{mdframed}

%% file: main.bbl
\begin{thebibliography}{59}
\providecommand{\natexlab}[1]{#1}

\bibitem[{Achiam et~al.(2023)Achiam, Adler, Agarwal, Ahmad, Akkaya, Aleman, Almeida, Altenschmidt, Altman, Anadkat et~al.}]{achiam2023gpt}
Josh Achiam, Steven Adler, Sandhini Agarwal, Lama Ahmad, Ilge Akkaya, Florencia~Leoni Aleman, Diogo Almeida, Janko Altenschmidt, Sam Altman, Shyamal Anadkat, et~al. 2023.
\newblock Gpt-4 technical report.
\newblock \emph{arXiv preprint arXiv:2303.08774}.

\bibitem[{Bai et~al.(2023)Bai, Bai, Chu, Cui, Dang, Deng, Fan, Ge, Han, Huang et~al.}]{bai2023qwen}
Jinze Bai, Shuai Bai, Yunfei Chu, Zeyu Cui, Kai Dang, Xiaodong Deng, Yang Fan, Wenbin Ge, Yu~Han, Fei Huang, et~al. 2023.
\newblock Qwen technical report.
\newblock \emph{arXiv preprint arXiv:2309.16609}.

\bibitem[{Beck(2020)}]{beck2020cognitive}
Judith~S Beck. 2020.
\newblock \emph{Cognitive behavior therapy: Basics and beyond}.
\newblock Guilford Publications.

\bibitem[{Brave and Nass(2007)}]{brave2007emotion}
Scott Brave and Cliff Nass. 2007.
\newblock Emotion in human-computer interaction.
\newblock In \emph{The human-computer interaction handbook}, pages 103--118. CRC Press.

\bibitem[{Castillo-Montoya(2016)}]{castillo2016preparing}
Milagros Castillo-Montoya. 2016.
\newblock Preparing for interview research: The interview protocol refinement framework.
\newblock \emph{Qualitative report}, 21(5).

\bibitem[{Chang et~al.(2023)Chang, Lo, Goyal, and Iyyer}]{Chang2023BooookScoreAS}
Yapei Chang, Kyle Lo, Tanya Goyal, and Mohit Iyyer. 2023.
\newblock Booookscore: A systematic exploration of book-length summarization in the era of llms.
\newblock \emph{ArXiv}.

\bibitem[{Chang et~al.(2024)Chang, Wang, Wang, Wu, Yang, Zhu, Chen, Yi, Wang, Wang et~al.}]{chang2024survey}
Yupeng Chang, Xu~Wang, Jindong Wang, Yuan Wu, Linyi Yang, Kaijie Zhu, Hao Chen, Xiaoyuan Yi, Cunxiang Wang, Yidong Wang, et~al. 2024.
\newblock A survey on evaluation of large language models.
\newblock \emph{ACM Transactions on Intelligent Systems and Technology}, 15(3):1--45.

\bibitem[{Chen et~al.(2024)Chen, Chen, Yan, Xu, Gao, Shen, Quan, Li, Zhang, Huang et~al.}]{chen2024roleinteract}
Hongzhan Chen, Hehong Chen, Ming Yan, Wenshen Xu, Xing Gao, Weizhou Shen, Xiaojun Quan, Chenliang Li, Ji~Zhang, Fei Huang, et~al. 2024.
\newblock Roleinteract: Evaluating the social interaction of role-playing agents.
\newblock \emph{arXiv preprint arXiv:2403.13679}.

\bibitem[{Clarke and Milne(2001)}]{clarke2001national}
Colin Clarke and Rebecca Milne. 2001.
\newblock \emph{A national evaluation of the PEACE Investigative Interviewing Course}.
\newblock Home office London.

\bibitem[{Cowie et~al.(2001)Cowie, Douglas-Cowie, Tsapatsoulis, Votsis, Kollias, Fellenz, and Taylor}]{cowie2001emotion}
Roddy Cowie, Ellen Douglas-Cowie, Nicolas Tsapatsoulis, George Votsis, Stefanos Kollias, Winfried Fellenz, and John~G Taylor. 2001.
\newblock Emotion recognition in human-computer interaction.
\newblock \emph{IEEE Signal processing magazine}, 18(1):32--80.

\bibitem[{Dai et~al.(2019)Dai, Yang, Yang, Carbonell, Le, and Salakhutdinov}]{dai2019transformer}
Zihang Dai, Zhilin Yang, Yiming Yang, Jaime Carbonell, Quoc~V Le, and Ruslan Salakhutdinov. 2019.
\newblock Transformer-xl: Attentive language models beyond a fixed-length context.
\newblock \emph{arXiv preprint arXiv:1901.02860}.

\bibitem[{Demszky et~al.(2023)Demszky, Yang, Yeager, Bryan, Clapper, Chandhok, Eichstaedt, Hecht, Jamieson, Johnson et~al.}]{demszky2023using}
Dorottya Demszky, Diyi Yang, David~S Yeager, Christopher~J Bryan, Margarett Clapper, Susannah Chandhok, Johannes~C Eichstaedt, Cameron Hecht, Jeremy Jamieson, Meghann Johnson, et~al. 2023.
\newblock Using large language models in psychology.
\newblock \emph{Nature Reviews Psychology}, 2(11):688--701.

\bibitem[{Douze et~al.(2024)Douze, Guzhva, Deng, Johnson, Szilvasy, Mazar{\'e}, Lomeli, Hosseini, and J{\'e}gou}]{douze2024faiss}
Matthijs Douze, Alexandr Guzhva, Chengqi Deng, Jeff Johnson, Gergely Szilvasy, Pierre-Emmanuel Mazar{\'e}, Maria Lomeli, Lucas Hosseini, and Herv{\'e} J{\'e}gou. 2024.
\newblock The faiss library.
\newblock \emph{arXiv preprint arXiv:2401.08281}.

\bibitem[{Duan et~al.(2022)Duan, Yu, Tan, Zhu, and Tan}]{duan2022survey}
Jiafei Duan, Samson Yu, Hui~Li Tan, Hongyuan Zhu, and Cheston Tan. 2022.
\newblock A survey of embodied ai: From simulators to research tasks.
\newblock \emph{IEEE Transactions on Emerging Topics in Computational Intelligence}, 6(2):230--244.

\bibitem[{Duan et~al.(2024{\natexlab{a}})Duan, Wang, Diffenderfer, Sun, Chen, Kailkhura, and Xu}]{duan-etal-2024-reta}
Jinhao Duan, Shiqi Wang, James Diffenderfer, Lichao Sun, Tianlong Chen, Bhavya Kailkhura, and Kaidi Xu. 2024{\natexlab{a}}.
\newblock \href {https://doi.org/10.18653/v1/2024.naacl-long.123} {{R}e{TA}: Recursively thinking ahead to improve the strategic reasoning of large language models}.
\newblock In \emph{Proceedings of the 2024 Conference of the North American Chapter of the Association for Computational Linguistics: Human Language Technologies (Volume 1: Long Papers)}, pages 2232--2246, Mexico City, Mexico. Association for Computational Linguistics.

\bibitem[{Duan et~al.(2024{\natexlab{b}})Duan, Zhang, Diffenderfer, Kailkhura, Sun, Stengel-Eskin, Bansal, Chen, and Xu}]{duan2024gtbench}
Jinhao Duan, Renming Zhang, James Diffenderfer, Bhavya Kailkhura, Lichao Sun, Elias Stengel-Eskin, Mohit Bansal, Tianlong Chen, and Kaidi Xu. 2024{\natexlab{b}}.
\newblock Gtbench: Uncovering the strategic reasoning limitations of llms via game-theoretic evaluations.
\newblock \emph{arXiv preprint arXiv:2402.12348}.

\bibitem[{Gao et~al.(2023)Gao, Xiong, Gao, Jia, Pan, Bi, Dai, Sun, and Wang}]{gao2023retrieval}
Yunfan Gao, Yun Xiong, Xinyu Gao, Kangxiang Jia, Jinliu Pan, Yuxi Bi, Yi~Dai, Jiawei Sun, and Haofen Wang. 2023.
\newblock Retrieval-augmented generation for large language models: A survey.
\newblock \emph{arXiv preprint arXiv:2312.10997}.

\bibitem[{Ham et~al.(2020)Ham, Lee, Jang, and Kim}]{ham-etal-2020-end}
Donghoon Ham, Jeong-Gwan Lee, Youngsoo Jang, and Kee-Eung Kim. 2020.
\newblock \href {https://doi.org/10.18653/v1/2020.acl-main.54} {End-to-end neural pipeline for goal-oriented dialogue systems using {GPT}-2}.
\newblock In \emph{Proceedings of the 58th Annual Meeting of the Association for Computational Linguistics}, pages 583--592, Online. Association for Computational Linguistics.

\bibitem[{Hettema et~al.(2005)Hettema, Steele, and Miller}]{hettema2005motivational}
Jennifer Hettema, Julie Steele, and William~R Miller. 2005.
\newblock Motivational interviewing.
\newblock \emph{Annu. Rev. Clin. Psychol.}, 1(1):91--111.

\bibitem[{Hong et~al.()Hong, Zheng, Meng, Liang, Chen, Dodge, Zhou, and Wang}]{hong2024conect}
Junyuan Hong, Wenqing Zheng, Han Meng, Siqi Liang, Anqing Chen, Hiroko~H Dodge, Jiayu Zhou, and Zhangyang Wang.
\newblock A-conect: Designing ai-based conversational chatbot for early dementia intervention.
\newblock In \emph{ICLR 2024 Workshop on Large Language Model (LLM) Agents}.

\bibitem[{Hosseini-Asl et~al.(2020)Hosseini-Asl, McCann, Wu, Yavuz, and Socher}]{NEURIPS2020_e9462095}
Ehsan Hosseini-Asl, Bryan McCann, Chien-Sheng Wu, Semih Yavuz, and Richard Socher. 2020.
\newblock \href {https://proceedings.neurips.cc/paper_files/paper/2020/file/e946209592563be0f01c844ab2170f0c-Paper.pdf} {A simple language model for task-oriented dialogue}.
\newblock In \emph{Advances in Neural Information Processing Systems}, volume~33, pages 20179--20191. Curran Associates, Inc.

\bibitem[{Huang et~al.(2023)Huang, Wang, Lam, Li, Jiao, and Lyu}]{huang2023psychobench}
Jen-tse Huang, Wenxuan Wang, M~Lam, E~Li, Wenxiang Jiao, and M~Lyu. 2023.
\newblock Revisiting the reliability of psychological scales on large language models.
\newblock \emph{arXiv preprint arXiv}, 2305.

\bibitem[{Inagaki et~al.(2023)Inagaki, Kato, Takahashi, Ozaki, and Kanda}]{Inagaki2023LLMsCG}
T~Inagaki, Akari Kato, Koichi Takahashi, Haruka Ozaki, and Genki~N. Kanda. 2023.
\newblock \href {https://api.semanticscholar.org/CorpusID:258236094} {Llms can generate robotic scripts from goal-oriented instructions in biological laboratory automation}.

\bibitem[{Jiang et~al.(2024)Jiang, Sablayrolles, Roux, Mensch, Savary, Bamford, Chaplot, Casas, Hanna, Bressand et~al.}]{jiang2024mixtral}
Albert~Q Jiang, Alexandre Sablayrolles, Antoine Roux, Arthur Mensch, Blanche Savary, Chris Bamford, Devendra~Singh Chaplot, Diego de~las Casas, Emma~Bou Hanna, Florian Bressand, et~al. 2024.
\newblock Mixtral of experts.
\newblock \emph{arXiv preprint arXiv:2401.04088}.

\bibitem[{Lamb et~al.(2007)Lamb, Orbach, Hershkowitz, Esplin, and Horowitz}]{lamb2007structured}
Michael~E Lamb, Yael Orbach, Irit Hershkowitz, Phillip~W Esplin, and Dvora Horowitz. 2007.
\newblock A structured forensic interview protocol improves the quality and informativeness of investigative interviews with children: A review of research using the nichd investigative interview protocol.
\newblock \emph{Child abuse \& neglect}, 31(11-12):1201--1231.

\bibitem[{Leichsenring and Leibing(2003)}]{leichsenring2003effectiveness}
Falk Leichsenring and Eric Leibing. 2003.
\newblock The effectiveness of psychodynamic therapy and cognitive behavior therapy in the treatment of personality disorders: A meta-analysis.
\newblock \emph{American journal of psychiatry}, 160(7):1223--1232.

\bibitem[{Li et~al.(2023{\natexlab{a}})Li, Leng, Yan, Shen, Wang, Mi, Fei, Feng, Yan, Wang et~al.}]{li2023chatharuhi}
Cheng Li, Ziang Leng, Chenxi Yan, Junyi Shen, Hao Wang, Weishi Mi, Yaying Fei, Xiaoyang Feng, Song Yan, HaoSheng Wang, et~al. 2023{\natexlab{a}}.
\newblock Chatharuhi: Reviving anime character in reality via large language model.
\newblock \emph{arXiv preprint arXiv:2308.09597}.

\bibitem[{Li et~al.(2023{\natexlab{b}})Li, Wang, Zhang, Zhu, Hou, Lian, Luo, Yang, and Xie}]{li2023large}
Cheng Li, Jindong Wang, Yixuan Zhang, Kaijie Zhu, Wenxin Hou, Jianxun Lian, Fang Luo, Qiang Yang, and Xing Xie. 2023{\natexlab{b}}.
\newblock Large language models understand and can be enhanced by emotional stimuli.
\newblock \emph{arXiv preprint arXiv:2307.11760}.

\bibitem[{Liu et~al.(2023)Liu, Lin, Hewitt, Paranjape, Bevilacqua, Petroni, and Liang}]{Liu2023LostIT}
Nelson~F. Liu, Kevin Lin, John Hewitt, Ashwin Paranjape, Michele Bevilacqua, Fabio Petroni, and Percy Liang. 2023.
\newblock \href {https://api.semanticscholar.org/CorpusID:259360665} {Lost in the middle: How language models use long contexts}.
\newblock \emph{Transactions of the Association for Computational Linguistics}, 12:157--173.

\bibitem[{Liu et~al.(2024)Liu, Yang, Zhang, Xie, Yu, and Ananiadou}]{liu2024emollms}
Zhiwei Liu, Kailai Yang, Tianlin Zhang, Qianqian Xie, Zeping Yu, and Sophia Ananiadou. 2024.
\newblock Emollms: A series of emotional large language models and annotation tools for comprehensive affective analysis.
\newblock \emph{arXiv preprint arXiv:2401.08508}.

\bibitem[{Luo et~al.(2023)Luo, Xie, and Ananiadou}]{luo2023chatgpt}
Zheheng Luo, Qianqian Xie, and Sophia Ananiadou. 2023.
\newblock Chatgpt as a factual inconsistency evaluator for abstractive text summarization.
\newblock \emph{arXiv preprint arXiv:2303.15621}.

\bibitem[{Maharana et~al.(2024)Maharana, Lee, Tulyakov, Bansal, Barbieri, and Fang}]{maharana2024lococmo}
Adyasha Maharana, Dong-Ho Lee, Sergey Tulyakov, Mohit Bansal, Francesco Barbieri, and Yuwei Fang. 2024.
\newblock Evaluating very long-term conversational memory of llm agents.
\newblock \emph{arxiv}.

\bibitem[{Maranto(2024)}]{maranto2024llmsat}
David Maranto. 2024.
\newblock Llmsat: A large language model-based goal-oriented agent for autonomous space exploration.
\newblock \emph{arXiv preprint arXiv:2405.01392}.

\bibitem[{Marcus(2018)}]{10.1093/actrade/9780199669240.001.0001}
Laura Marcus. 2018.
\newblock \href {https://doi.org/10.1093/actrade/9780199669240.001.0001} {\emph{{Autobiography: A Very Short Introduction}}}.
\newblock Oxford University Press.

\bibitem[{Maunsell(2016)}]{2016maunsell}
Jerome Maunsell. 2016.
\newblock \href {https://doi.org/10.5463/ejlw.5.194} {"the literary interview as autobiography"}.
\newblock \emph{European Journal of Life Writing}, 5:MC23.

\bibitem[{McAdams(2008)}]{mcadams2008life}
Dan~P McAdams. 2008.
\newblock The life story interview.

\bibitem[{Mehri et~al.(2020)Mehri, Eric, and Hakkani-Tur}]{MehriDialoGLUE2020}
S.~Mehri, M.~Eric, and D.~Hakkani-Tur. 2020.
\newblock Dialoglue: A natural language understanding benchmark for task-oriented dialogue.
\newblock \emph{ArXiv}, abs/2009.13570.

\bibitem[{Meta(2024)}]{llama3}
Meta. 2024.
\newblock Introducing meta llama 3: The most capable openly available llm to date.
\newblock \url{https://ai.meta.com/blog/meta-llama-3/}.
\newblock Accessed: 2024-05-18.

\bibitem[{Nye et~al.(2023)Nye, Mee, and Core}]{nye2023generative}
B~Nye, Dillon Mee, and Mark~G Core. 2023.
\newblock Generative large language models for dialog-based tutoring: An early consideration of opportunities and concerns.
\newblock In \emph{AIED Workshops}.

\bibitem[{Ouyang et~al.(2022)Ouyang, Wu, Jiang, Almeida, Wainwright, Mishkin, Zhang, Agarwal, Slama, Ray et~al.}]{ouyang2022training}
Long Ouyang, Jeffrey Wu, Xu~Jiang, Diogo Almeida, Carroll Wainwright, Pamela Mishkin, Chong Zhang, Sandhini Agarwal, Katarina Slama, Alex Ray, et~al. 2022.
\newblock Training language models to follow instructions with human feedback.
\newblock \emph{Advances in neural information processing systems}, 35:27730--27744.

\bibitem[{Panagoulias et~al.(2024)Panagoulias, Virvou, and Tsihrintzis}]{panagoulias2024augmenting}
Dimitrios~P Panagoulias, Maria Virvou, and George~A Tsihrintzis. 2024.
\newblock Augmenting large language models with rules for enhanced domain-specific interactions: The case of medical diagnosis.
\newblock \emph{Electronics}, 13(2):320.

\bibitem[{Pasupathi et~al.(2007)Pasupathi, Mansour, and Brubaker}]{10.1159/000100939}
M.~Pasupathi, E.~Mansour, and J.R. Brubaker. 2007.
\newblock \href {https://doi.org/10.1159/000100939} {{Developing a Life Story: Constructing Relations between Self and Experience in Autobiographical Narratives}}.
\newblock \emph{Human Development}, 50(2-3):85--110.

\bibitem[{Quora(2021)}]{quoraRateCritique}
Quora. 2021.
\newblock {H}ow do you rate and critique memoirs and autobiographies? --- quora.com.
\newblock https://www.quora.com/How-do-you-rate-and-critique-memoirs-and-autobiographies.

\bibitem[{Rautalinko et~al.(2007)Rautalinko, Lisper, and Ekehammar}]{rautalinko2007reflective}
Erik Rautalinko, Hans-Olof Lisper, and Bo~Ekehammar. 2007.
\newblock Reflective listening in counseling: effects of training time and evaluator social skills.
\newblock \emph{American journal of psychotherapy}, 61(2):191--209.

\bibitem[{Safdari et~al.(2023)Safdari, Serapio-Garc{\'\i}a, Crepy, Fitz, Romero, Sun, Abdulhai, Faust, and Matari{\'c}}]{safdari2023personality}
Mustafa Safdari, Greg Serapio-Garc{\'\i}a, Cl{\'e}ment Crepy, Stephen Fitz, Peter Romero, Luning Sun, Marwa Abdulhai, Aleksandra Faust, and Maja Matari{\'c}. 2023.
\newblock Personality traits in large language models.
\newblock \emph{arXiv preprint arXiv:2307.00184}.

\bibitem[{Smorti(2011)}]{jbp:/content/journals/10.1075/ni.21.2.08smo}
Andrea Smorti. 2011.
\newblock \href {https://doi.org/10.1075/ni.21.2.08smo} {Autobiographical memory and autobiographical narrative: What is the relationship?}
\newblock \emph{Narrative Inquiry}, 21(2):303--310.

\bibitem[{Tao et~al.(2023)Tao, Liang, Shi, Yu, and Xie}]{tao2023rolecraft}
Meiling Tao, Xuechen Liang, Tianyu Shi, Lei Yu, and Yiting Xie. 2023.
\newblock Rolecraft-glm: Advancing personalized role-playing in large language models.
\newblock \emph{arXiv preprint arXiv:2401.09432}.

\bibitem[{Tao et~al.(2024)Tao, Agrawal, Dombi, Sydorenko, and Lee}]{tao2024chatgpt}
Yufei Tao, Ameeta Agrawal, Judit Dombi, Tetyana Sydorenko, and Jung~In Lee. 2024.
\newblock Chatgpt role-play dataset: Analysis of user motives and model naturalness.
\newblock \emph{arXiv preprint arXiv:2403.18121}.

\bibitem[{Wang et~al.(2024{\natexlab{a}})Wang, Ma, Feng, Zhang, Yang, Zhang, Chen, Tang, Chen, Lin et~al.}]{wang2024survey}
Lei Wang, Chen Ma, Xueyang Feng, Zeyu Zhang, Hao Yang, Jingsen Zhang, Zhiyuan Chen, Jiakai Tang, Xu~Chen, Yankai Lin, et~al. 2024{\natexlab{a}}.
\newblock A survey on large language model based autonomous agents.
\newblock \emph{Frontiers of Computer Science}, 18(6):186345.

\bibitem[{Wang et~al.(2023{\natexlab{a}})Wang, ping Peng, Zha, Han, Deng, Hu, and Hu}]{Wang2023EnhancingTC}
Qing Wang, Shu ping Peng, Zhiyuan Zha, Xue Han, Chao Deng, Lun Hu, and Pengwei Hu. 2023{\natexlab{a}}.
\newblock Enhancing the conversational agent with an emotional support system for mental health digital therapeutics.
\newblock \emph{Frontiers in Psychiatry}, 14.

\bibitem[{Wang et~al.(2024{\natexlab{b}})Wang, Milani, Chiu, Eack, Labrum, Murphy, Jones, Hardy, Shen, Fang et~al.}]{wang2024patient}
Ruiyi Wang, Stephanie Milani, Jamie~C Chiu, Shaun~M Eack, Travis Labrum, Samuel~M Murphy, Nev Jones, Kate Hardy, Hong Shen, Fei Fang, et~al. 2024{\natexlab{b}}.
\newblock Patient-$\{$$\backslash$Psi$\}$: Using large language models to simulate patients for training mental health professionals.
\newblock \emph{arXiv preprint arXiv:2405.19660}.

\bibitem[{Wang et~al.(2023{\natexlab{b}})Wang, Xiao, tse Huang, Yuan, Xu, Guo, Tu, Fei, Leng, Wang et~al.}]{wang2024incharacter}
Xintao Wang, Yunze Xiao, Jen tse Huang, Siyu Yuan, Rui Xu, Haoran Guo, Quan Tu, Yaying Fei, Ziang Leng, Wei Wang, et~al. 2023{\natexlab{b}}.
\newblock Incharacter: Evaluating personality fidelity in role-playing agents through psychological interviews.
\newblock \emph{arXiv preprint arXiv:2310.17976}.

\bibitem[{Wang et~al.(2023{\natexlab{c}})Wang, Peng, Que, Liu, Zhou, Wu, Guo, Gan, Ni, Zhang et~al.}]{wang2023rolellm}
Zekun~Moore Wang, Zhongyuan Peng, Haoran Que, Jiaheng Liu, Wangchunshu Zhou, Yuhan Wu, Hongcheng Guo, Ruitong Gan, Zehao Ni, Man Zhang, et~al. 2023{\natexlab{c}}.
\newblock Rolellm: Benchmarking, eliciting, and enhancing role-playing abilities of large language models.
\newblock \emph{arXiv preprint arXiv:2310.00746}.

\bibitem[{Wu et~al.(2020)Wu, Hoi, Socher, and Xiong}]{wu-etal-2020-tod}
Chien-Sheng Wu, Steven~C.H. Hoi, Richard Socher, and Caiming Xiong. 2020.
\newblock \href {https://doi.org/10.18653/v1/2020.emnlp-main.66} {{TOD}-{BERT}: Pre-trained natural language understanding for task-oriented dialogue}.
\newblock In \emph{Proceedings of the 2020 Conference on Empirical Methods in Natural Language Processing (EMNLP)}, pages 917--929, Online. Association for Computational Linguistics.

\bibitem[{Wu et~al.(2023)Wu, Bansal, Zhang, Wu, Zhang, Zhu, Li, Jiang, Zhang, and Wang}]{wu2023autogen}
Qingyun Wu, Gagan Bansal, Jieyu Zhang, Yiran Wu, Shaokun Zhang, Erkang Zhu, Beibin Li, Li~Jiang, Xiaoyun Zhang, and Chi Wang. 2023.
\newblock Autogen: Enabling next-gen llm applications via multi-agent conversation framework.
\newblock \emph{arXiv preprint arXiv:2308.08155}.

\bibitem[{Xu et~al.(2022)Xu, Zhang, Wu, and Wei}]{xu2022sequence}
Shusheng Xu, Xingxing Zhang, Yi~Wu, and Furu Wei. 2022.
\newblock Sequence level contrastive learning for text summarization.
\newblock In \emph{Proceedings of the AAAI conference on artificial intelligence}, volume~36, pages 11556--11565.

\bibitem[{Zhang et~al.(2021)Zhang, Ni, Mao, Wu, Zhu, Deb, Awadallah, Radev, and Zhang}]{zhang2021summ}
Yusen Zhang, Ansong Ni, Ziming Mao, Chen~Henry Wu, Chenguang Zhu, Budhaditya Deb, Ahmed~H Awadallah, Dragomir Radev, and Rui Zhang. 2021.
\newblock Summ\^{} n: A multi-stage summarization framework for long input dialogues and documents.
\newblock \emph{arXiv preprint arXiv:2110.10150}.

\bibitem[{Zheng et~al.(2024)Zheng, Chiang, Sheng, Zhuang, Wu, Zhuang, Lin, Li, Li, Xing et~al.}]{zheng2024judging}
Lianmin Zheng, Wei-Lin Chiang, Ying Sheng, Siyuan Zhuang, Zhanghao Wu, Yonghao Zhuang, Zi~Lin, Zhuohan Li, Dacheng Li, Eric Xing, et~al. 2024.
\newblock Judging llm-as-a-judge with mt-bench and chatbot arena.
\newblock \emph{Advances in Neural Information Processing Systems}, 36.

\bibitem[{Zhong et~al.(2022)Zhong, Liu, Xu, Zhu, and Zeng}]{zhong2022dialoglm}
Ming Zhong, Yang Liu, Yichong Xu, Chenguang Zhu, and Michael Zeng. 2022.
\newblock Dialoglm: Pre-trained model for long dialogue understanding and summarization.
\newblock In \emph{Proceedings of the AAAI Conference on Artificial Intelligence}, volume~36, pages 11765--11773.

\end{thebibliography}
